\documentclass[12pt]{article}%
\usepackage[switch,columnwise]{lineno}
\usepackage{amsmath}%
\usepackage[dvips]{color}
\usepackage{txfonts}%
\usepackage{amsfonts}%
\usepackage{amssymb}%
\usepackage{graphicx}
\usepackage{color}
\usepackage{setspace}
\usepackage{esint}
\usepackage[rightcaption]{sidecap}
\usepackage{wrapfig}
\usepackage{lipsum}
\usepackage{caption}
\usepackage{longtable}
\usepackage{multirow}
\usepackage[T2A,T1]{fontenc}
\usepackage[utf8]{inputenc}
\usepackage[russian,english]{babel}

\numberwithin{equation}{section}

\newcommand{\be}{\begin{equation}}
\newcommand{\ee}{\end{equation}}

\usepackage{anysize}
\marginsize{2.5cm}{2.5cm}{1cm}{2cm}
\usepackage{mathptmx}
-1in\relax 

\begin{document}

\begin{center}
\LARGE
\textbf{Drone Launched Short Range Rockets}
\normalsize
\vskip1cm
\begin{tabular}{ll}
\Large
\Large Mikhail V. Shubov         \\
\normalsize
University of MA Lowell    \\
One University Ave,        \\
Lowell, MA 01854           \\
E-mail: viktor\_shubov@uml.edu  \\
\end{tabular}
\end{center}

\begin{center}
\textbf{Abstract}
\end{center}
\begin{quote}\begin{quote}
\hskip.5cm A  concept of drone launched short range rockets (DLSRR) is presented.  A drone or an aircraft rises DLSRR to a release altitude of up to 20 $km$.  At the release altitude, the drone or an aircraft is moving at a velocity of up to 700 $m/s$ and a steep angle of up to 68$^o$ to the horizontal.  After DLSRRs are released, their motors start firing.  DLSRRs use slow burning motors to gain altitude and velocity.  At the apogee of their flight DLSRRs release projectiles which fly to the target and strike it at high impact velocity.  The projectiles reach a target at ranges of up to 442 $km$ and impact velocities up to 1.88 $km/s$.  We show that a rocket launched at high altitude and high initial velocity does not need expensive thermal protection to survive ascent.  Delivery of munitions to target by DLSRRs should be much less expensive than delivery by a conventional rocket.
Even though delivery of munitions by bomber aircraft is even less expensive, a bomber needs to fly close to the target, while a DLSRR carrier releases the rockets from a distance of at least 200 $km$ from the target.
All parameters of DLSRRs, and their trajectories are calculated based on theoretical (mechanical and thermodynamical) analysis and on several MatLab programs.
\end{quote}\end{quote}

\begin{center}
  \textbf{List of Abbreviations -- General}
\end{center}
\emph{ATACMS}  -- army tactical missile system (USA)\\
\emph{CEP}     -- circular error probability\\
\emph{DF}      -- Dong-Feng ballistic missiles (China)\\
\emph{DLSRR}   -- drone launched short range rockets\\
\emph{GMLRS}   -- guided multiple launch rocket system (USA)\\
\emph{HPV}     -- hypervelocity projectile\\
\emph{JDAM}    -- joint direct attack munition (guidance kit for any bomb)\\
\emph{LORA}    -- long range attack missile (Israel)\\
\emph{MKV}     -- multiple kill vehicle\\
\emph{MLRS}    -- multiple launch rocket system\\
\emph{PGK}     -- precision guidance kit\\
\emph{RASAero} -- aerodynamic analysis and flight simulation software\\
\emph{RLA}     -- rocket launcher aircraft\\
\emph{RPA}     -- rocket propulsion analysis\\
\emph{SAM}     -- surface to air missile\\
\emph{SRBM}    -- short range ballistic missiles\\
\emph{TBM}     -- theatre ballistic missile\\

\begin{center}
  \textbf{List of Abbreviations -- Chemicals}
\end{center}
\emph{AP}    -- ammonium perchlorate\\
\emph{AN}    -- ammonium nitrate\\
\emph{HTPB}  -- hydroxyl-terminated polybutadiene\\

\Large
\begin{center}
\textbf{Introduction}
\end{center}
\normalsize
In this work we discuss a possibility of a short range rocket launched from a fighter or a drone.  The carrier aircraft or drone is called Rocket Launcher Aircraft (RLA).  RLA can either be a modified fighter like F16, or an aircraft or drone constructed specifically for a rocket-launching mission.  The rocket or rockets are called  Drone Launched Short Range Rockets (DLSRR).  In this work, we present the models and calculate performance of DLSRR, but not RLA.  The author plans to write another work on RLA.

RLA raises DLSRRs to an altitude of 12 $km$ -- 20 $km$ and fires them.  At the firing point, RLA and hence DLSRRs have a velocity of 500 $m/s$ -- 700 $m/s$ in the ground system of coordinates.  Their motion is at an angle of 53$^o$ to 68$^o$ with respect to the horizontal.  DLSRR has a slow-burning motor which burns 60 $s$ -- 80 $s$.  After the DLSRR motor burns out, it follows a ballistic trajectory.  As the DLSRR reaches the apogee of 77 $km$ -- 111 $km$, it releases one or more warheads which fly to the target.  The DLSRR itself lands a few $km$ away from the target.

Very little work has been done on military rockets with reusable first stage.
Between 2010 and 2012, US Air Force had a project called ``Reusable Booster System" to assist long-range rockets \cite{rbs1}.  
Jeffery Becker's concept is Theater Guided Missile Carrier (T-CVG), which is similar to our RLA.  In Becker's concept, a modified version of reusable first stage of Falcon I rocket is used as RLA.  This stage rises almost vertically, firing the second stage rockets which have a range of 1,600 $km$ to 8,000 $km$ \cite{cgmc}.  Our proposal is more modest.  It is likely, that multiple forms of military rocket systems with reusable first stage will exist.

Delivery of payload to target should be less expensive than conventional short range rocket systems.  It may be noticed that conventional fighters and bombers also deliver munitions to a relatively long distance at a relatively low price.  Unlike RLA, conventional fighters and bombers expose themselves to antiaircraft fire, which may make their operations inefficient.  RLA never flies into hostile territory.

The current work consists of 6 Sections.  In Section 1 we describe the state of art technology for long range precision strikes.  In Subsection 1.1, we describe the state of the art rocket systems capable of making precise strikes to the range of 80 $km$ to 600 $km$ also mentioning some longer range missiles.  In Tables \ref{T01} and \ref{T02}, we present the cost of delivery per $kg$ of payload.  For ranges of 156 $km$ to 600 $km$, the costs are from \$ 2,700 per $kg$ to \$ 6,700 per $kg$ depending on the missile.  Thus we establish the need for a less expensive system.
In Subsection 1.2, we describe historic and state of art guided munitions.
In Subsection 1.3, we describe some antiaircraft missiles.  The main reason the RLA -- DLSRR system has not been developed so far is that bomber aircraft had capability of long range precision strike.  The new generation of antiaircraft missiles may make bombers obsolete in any sizable confrontation.

In Section 2, we introduce the physics of rocket flight.  We present the equations describing the rocket trajectory.  We also introduce the concepts of Specific Impulse, gravity loss, and drag loss.  These concepts are important for rocket development.  The material in Section 2 is used to write programs and perform calculations with results presented in subsequent sections.

In Section 3, we discuss solid propellant rockets.
In Subsection 3.1, we describe the rockets we need for DLSRRs
In Subsection 3.2, we present properties of several fuel compositions.
In Subsection 3.3, we discuss possible fuel grain shapes.  We also discuss different rocket container materials along with their corresponding advantages and disadvantages.

In Section 4, we calculate rocket and warhead performance.
In Subsection 4.1 we describe the performance of DLSRRs fired from RLA.
Ranges obtained for these rockets are 263 $km$ to 442 $km$ depending on launch conditions and propellant burning time.
In Subsection 4.2 we calculate the impact velocity of several projectiles.
The Hypervelocity Projectile is described in Subsection 4.4.1.
Europrojectile is a small tungsten projectile described in Subsubsection 4.2.2.
Europrojectile has the best performance.

Fired from 12 $km$ altitude with an initial velocity of 500 $m/s$, and rocket burning time of 80 $s$, DLSRR has a range of 264 $km$.
In this case the impact velocity of the Europrojectile is 1,400 $m/s$.
Fired from 16 $km$ altitude with an initial velocity of 600 $m/s$, and rocket burning time of 50 $s$, DLSRR has a range of 361 $km$.
In this case the impact velocity of the Europrojectile is 1,675 $m/s$.
Fired from 20 $km$ altitude with an initial velocity of 700 $m/s$, and rocket burning time of 30 $s$, DLSRR has a range of 442 $km$.
In this case the impact velocity of the Europrojectile is 1,879 $m/s$.
High release altitude, high release velocity, and short propellant burning times enhance rocket range and warhead impact velocity.  Higher release altitude and velocity place greater requirement on RLA, thus making the launch of each DLSRR more expensive.

In Section 5, we describe the aerodynamic heating and thermal protection.
In Subsection 5.1 we present the general expressions for aerodynamic heating of different parts of the rocket.
In Subsubsection 5.1.1, we present the equations for the stagnation point heat flux.
In Subsubsection 5.1.2, we present the equations for general turbulent flow heat flux.
In Subsubsection 5.1.3, we present the equations for the heat flux on rocket cylinder.
In Subsubsection 5.1.4, we present the equations for the nose cone heat flux.
The material in Subsection 5.1 is used to write programs and perform calculations with the results presented in Subsection 5.3.
In Subsection 5.2 we discuss the heat shields.
In Subsubsection 5.2.1, we describe radiation heat shields.
In Subsubsection 5.2.2, we describe heat sink shields.
In Subsection 5.3 we calculate the heating rates and temperatures of different parts of DLSRR during flight.
In Subsubsection 5.3.1, we calculate stagnation point temperature as a function of time.  The temperature never exceeds 1,200 $^o$C -- well within the capability of thermal protection.
In Subsubsection 5.3.2, we calculate temperatures of several points on the nose cone as a function of time.  The temperature never exceeds 960 $^o$C.
In Subsubsection 5.3.2, we calculate temperature DLSRR cylinder base as a function of time.  The cylinder base temperature never exceeds 94 $^o$C -- at this temperature, aluminum does not soften.

In Section 6, we present a conclusion.  In Appendix, we describe the MatLab programs we have written and used in this work.


\section{Artillery Rockets and Warheads}
\subsection{Artillery Rockets}
Short range ballistic missiles can make precise strikes at ranges from 150 $km$ to 1,000 $km$.  Some of the modern systems are extremely expensive.  As we see from Table \ref{T01} and Table \ref{T02}, the cost per 1 $kg$ of munitions delivered to the target varies between \$1,500 and \$6,500 depending on the range.

Scud missile was the original Soviet short range ballistic missile.
Scud-A introduced in 1957 delivered a 950 $kg$ payload to a range of 180 $km$.  Scud-B introduced in 1964 delivered a 985 $kg$ payload to a range of 300 $km$.  Over 7,000 Scud-B missiles have been built.
Scud-C introduced in 1965 delivered a 600 $kg$ payload to a range of 600 $km$.
Scud-D introduced in 1989 delivered a 985 $kg$ payload to a range of 700 $km$.
Scud missiles have been used extensively in \emph{Iran-Iraq War}.  Between October 1988 and February 1992 approximately 1,700 to 2,000 Scud missiles were launched in Afghanistan's civil war \cite{scud}.
Original Scud missiles had poor guidance, but in modern times, relatively inexpensive and precise guidance systems are available.
Russian short range rocket Iskander delivers a 480 $kg$ payload to a distance of 500 $km$ or a 700 $kg$ unit to a distance of 400 $km$.  The missile has very precise guidance.  Russia has only 136 such missiles.  The price is kept secret, but should be at least \$4 Million per unit \cite[p.125]{iskander}.

American short range rocket ATACMS delivers a 230 $kg$ payload to 300 $km$ or a 560 $kg$ payload to 165 $km$.  The missile has very precise guidance. USA has 3,700 units \cite{atacms}.  The unit cost of ATACMS is \$1.5 Million \cite[p.19]{defense16}.
Israel has developed LORA ("Long Range Attack") missile, which delivers a 570 $kg$ payload to a distance of 400 $km$.  The missile has very precise guidance.  The unit cost is not disclosed \cite{lora}.

China has Dong-Feng ballistic missiles.  Some of these missiles are intercontinental ballistic missiles.  Main short-range ballistic missiles are DF-11 and DF-15.
DF-11  launches an 800 $kg$ payload to a range of 280 $km$ or a 500 $kg$ payload to a range of 350 $km$.
DF-11A launches an 500 $kg$ payload to a range of 350 $km$.
DF-15 launches an 500 $kg$ payload to a range of 600 $km$.
DF-15A launches an 600 $kg$ payload to a range of 600 $km$.
DF-15B launches an 600 $kg$ payload to a range of 700 $km$ \cite[p.48]{DF11}.
China possesses about 1,200 ballistic rockets with ranges 300 $km$ to 1,000 $km$ \cite[p.31]{china}.

Artillery rockets can deliver guided submunitions much better than artillery projectiles, since they are not subject to extreme accelerations.
Guided Multiple Launch Rocket System (GMLRS) is used by USA.  M30 and M31 rockets deliver a 90 $kg$ payload to 84 $km$. The missile has very precise guidance \cite[p.112]{gmlrs}.  By 2014, 19,942 GMLRS rockets have been procured at a cost of \$125,850 per rocket.  During the years 2014-2015, 2,940 GMLRS rockets have been procured at a cost of \$400 Million \cite[p.19]{defense16} -- which gives GMLRS rocket a unit price of \$136,000.  Smerch is a powerful Russian rocket artillery system.  A single vehicle carries 12 rockets.  A single rocket delivers a 243 $kg$ warhead to a range of 90 $km$ with great accuracy.  Multiple Launch Rocket System (MLRS) Smerch has been used in \emph{Second Chechen War}, \emph{War in Donbass}, and \emph{Syrian Civil War}.  The unit price is undisclosed \cite{smerch}.  China has several powerful rocket artillery systems.  AR3 is a 370 $mm$ caliber rocket which delivers a 200 $kg$ warhead to a range of 220 $km$ with great accuracy.  Each launcher has 8 rockets \cite{AR3}.

USA did use the rockets described above in recent local conflicts.  Out of 17,184 GMLRS Unitary rockets produced, 3,141 have been used in Iraq and Afghanistan.  Out of 1,650 Army Tactical Missile System (ATACMS) Block I rockets produced, 32 have been used in \emph{Desert Storm} and 379 in \emph{Operation Iraqi Freedom}.
Out of 610 ATACMS Block IA rockets produced, 74 have been used in \emph{Operation Iraqi Freedom}.
Out of 176 ATACMS Quick Reaction Unitary rockets produced, 16 have been used in \emph{Operation Iraqi Freedom} and 42 in \emph{Operation Enduring Freedom}.
Out of 513 ATACMS 2000 rockets produced, 33 have been used in \emph{Operation Enduring Freedom} \cite{ruse}.

The measure of expense for a missile system at a given range is the cost per unit delivered weight.  Below we tabulate the costs for the few solid propellant rockets for which the information is not classified:
\begin{center}
\begin{tabular}{|l|l|l|l|l|}
  \hline
  Missile & Range    & Throw   & Unit      & Cost per $kg$ \\
          &          & weight  & Cost      & delivered\\
  \hline
  GMLRS   & 84 $km$  & 90  $kg$ & \$ 136,000   & \$ 1,500 \\
  ATACMS  & 165 $km$ & 560 $kg$ & \$ 1,500,000 & \$ 2,700 \\
  ATACMS  & 300 $km$ & 230 $kg$ & \$ 1,500,000 & \$ 6,500 \\
  Trident II \cite{trident} &7,840 $km$ & 2,800 $kg$ & \$ 37,300,000 & \$ 13,300\\
  \hline
\end{tabular}
\captionof{table}{Solid propellant missiles \label{T01}}
\end{center}
Liquid propellant rockets are less expensive, but they require about an hour to be prepared for firing.  Some of them are listed in Table \ref{T02} below.  The first two prices in 1990s are given in \cite{LiqRock}.  The prices listed below are double the old prices.
\begin{center}
\begin{tabular}{|l|l|l|l|l|}
  \hline
  Missile & Range    & Throw   & Unit      & Cost per $kg$ \\
          &          & weight  & Cost      & delivered\\
  \hline
  Dong-Feng 15        &   500 $km$ &   600 $kg$ & \$ 4,000,000 & \$ 6,700 \\
  Scud C              &   600 $km$ &   600 $kg$ & \$ 2,000,000 & \$ 3,300 \\
  Agni II \cite{Agni} & 2,000 $km$ & 1,000 $kg$ & \$ 5,100,000 & \$ 5,100 \\
  \hline
\end{tabular}
\captionof{table}{Liquid propellant missiles \label{T02}}
\end{center}

As can be seen from the above table, the cost of delivering munitions by short range ballistic missiles is very high.  The costs of long range missiles are relatively low for their range.

\subsection{Guided Munitions}
One of the most important factors enabling long-range missiles and artillery is appearance and development of guided munitions.  Precise guidance allows projectiles to strike targets tens or hundreds kilometers away with an accuracy of a few meters.

The most common guided munition is the air bomb. Guided bombs and glide bombs were first used by Germany in 1943 \cite[p.4]{Watts02}.  Since that time, relatively inexpensive guided bombs came into common use -- during operations \emph{Enduring Freedom} in 2001 and \emph{Iraqi Freedom} in 2003, over 50\% of all bombs used were guided \cite[p. 20]{Watts02}.  Joint Direct Attack Munition (JDAM) is a guidance kit which can be attached to any unguided bomb in order to transfer it into a guided bomb \cite[p.213]{Watts02}.  "Through FY 2005, 105,286 JDAM kits had been procured for an average unit-procurement cost of \$21,379 each" \cite[p.217]{Watts02}.  Recently the Boeing corporation has produced 300,000$^{\text{th}}$ JDAM \cite{jdam}.  Warheads used by the system we are proposing in this paper are similar to bombs dropped from an apogee of the second-stage rocket flight.  They are much lighter than conventional bombs, but they will experience strong heating due to high velocity.

Guided projectiles and projectile guidance kits are much more hardy than bomb or missile guidance -- they must survive acceleration of thousands of g's inside a cannon.
\begin{wrapfigure}{R}{12cm}
\vskip-.1cm
\includegraphics[width=12cm,height=6cm]{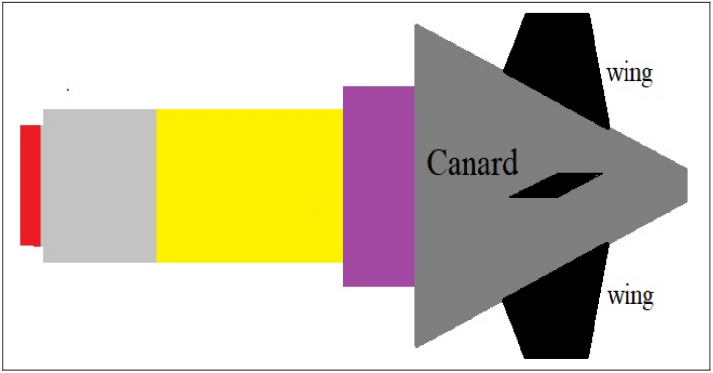}
\vskip-.8cm
\captionof{figure}{Projectile guidance kit \label{F01}}
\end{wrapfigure}
Low cost guidance systems for cannon launched projectiles are becoming available \cite{lcg01}.  The present generation of guided tube artillery munitions uses precision guidance kits attached to non-guided projectiles.  The kit currently used by US Army is "M1156 Precision Guidance Kit (PGK)".  PGK consists of a fuse placed in the nose of a non-guided projectile.  M1156 uses GPS guidance.  The projectile equipped with a fuse was suppose to have a circular error probability (CEP) of 30 $m$ \cite[p. 21]{Armada}.  Almost always new technology performs below expected parameters with M1156 being a nice exception.  ``During its first lot acceptance testing in April 2015, the M1156 demonstrated median accuracy of less than 10 $m$ with a reliability of 97\% when fired from the 155 $mm$/39-calibre M109A6 Paladin."  \cite[p.8]{PGK2}  A German test had even better result. "From a distance of 27 $km$, 90 percent of the PGK-equipped German shells landed within 5 meters of the target" \cite{PGK001}.  PGK M1156 have a unit price of \$10,000 \cite{PGK002}.

In 2013, 1,300 PGKs have been delivered to US Armed Forces in Afghanistan.   Australia has a contract with USA for 4,000 PGKs \cite[p. 21]{Armada}.
By May 2018, 25,000 units have been produced \cite{Orbital1}.
One of the main advantages of M1156 is its simplicity.  Indeed, (see Fig. \ref{F01})
\begin{quote}
   The kit has a single moving part, the canard assembly,which can only rotate along the longitudinal axis, the wings having a fixed cant; two couples of opposite wings have the same direction and thus provide lift, while the two despin wings provide counter-rotation \cite[p. 21]{Armada}.
\end{quote}
100,000 PGKs are ordered \cite[p.8]{PGK2}.   Israel is also developing a guidance kit called "Silver Bullet".  The kit can be fitted to the nose of any conventional or rocket assisted projectile.  This PGK does have four moving wings \cite[p. 21-22]{Armada}.  Another guidance kit developed by Israel is TopGun \cite{TopGun}.  Guidance kits used to guide projectiles should be able to guide warheads released by second-stage rockets.

The Hypervelocity Projectile (HPV) has been produced and tested.  It is likely the next generation of guided munitions \cite[p.12]{HVP01}.  Its guidance should easily survive extreme acceleration within a launcher and heating within atmosphere.  The projectile's guidance system should work even if it is shot from an electromagnetic cannon to a range of 400 $km$.  HPV has a conical shape.   It is made mostly out of tungsten.  HPV's weight is 11.4 $kg$, length 64.2 $cm$, and caliber 7.83 $cm$.  It is shown in Figure \ref{F02} below:
\begin{center}
\hspace*{-2.5cm}\includegraphics[width=20cm,height=4.5cm]{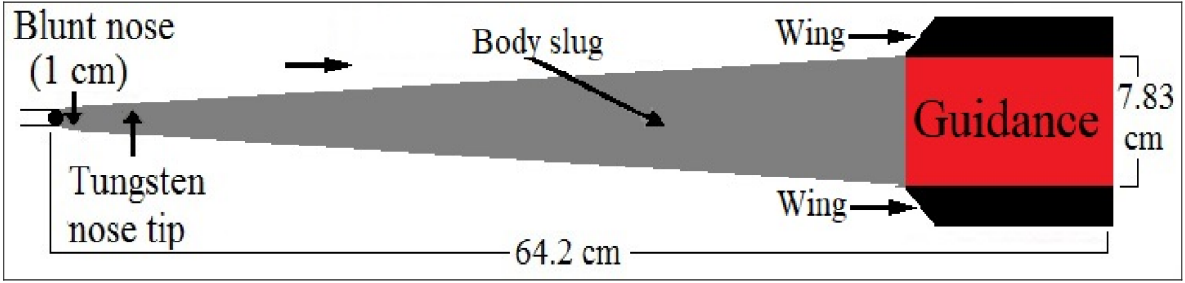}
\vskip-.6cm
\captionof{figure}{Hypervelocity projectile \label{F02}}
\vskip.4cm
\end{center}
HPV is very fit to be one of possible warheads carried by a drone launched short range rocket.

\subsection{Antiaircraft Missiles}

One of the main reasons the DLSRR technology has not been developed this far is that technology for precision strike at long range existed for many decades.  As we have discussed in Subsection 1.2, bomber aircraft made it possible to bring guided munition to the target at a relatively low cost.  Thus, there was no need for any unconventional delivery system.

In large-scale modern warfare, bomber aircraft may become a lost technology due to proliferation of very powerful anti-aircraft missiles.  The main Russian surface to air missile in 2020 is S400.  S400 missile with GRAU index 40N6 has a range of 400 $km$, ceiling of 185 $km$, and maximum velocity 2,000 $m/s$.  The unit cost of the missile is \$3.6 Million \cite{sam_s400}.  A new SAM called S500 should exceed the capabilities of S400.  A new generation of SAMs may also employ Multiple Kill Vehicle (MKV).  One rocket releases several guided submunitions each of which is capable of destroying the targeted aircraft \cite[p.42]{sam_mkv}.

USA also has very powerful antiaircraft missiles.  SM3 Block 1B missile can destroy any aircraft or an incoming missile within 700 $km$.  The warhead weighs 29 $kg$.  The missile costs \$12 million in 2018 dollars \cite{SM3}.   SM6 missile can destroy any aircraft within 300 $km$, it has a 64 $kg$ warhead and a price tag of \$4 million in 2018 dollars \cite{SM6}.

These missiles would make it impossible for any conventional bomber to operate in any sizable conflict.  RLA does not approach the target closer then about 200 $km$.  Moreover, it is possible to design an RLA which stays in the air only for a short period of time not exceeding 10 minutes.

\section{The Physics of Rocket Flight}

Consider a rocket that starts ascent at an altitude $h_0$, which is generally 12 $km$ to 20 $km$, with the initial speed is $v_0$, which is generally 500 $m/s$ to 700 $m/s$, with the  angle with respect to horizontal line $\alpha_0$, generally between 51$^o$ and 68$^o$.

\begin{wrapfigure}{R}{8cm}
\includegraphics[width=8cm,height=8cm]{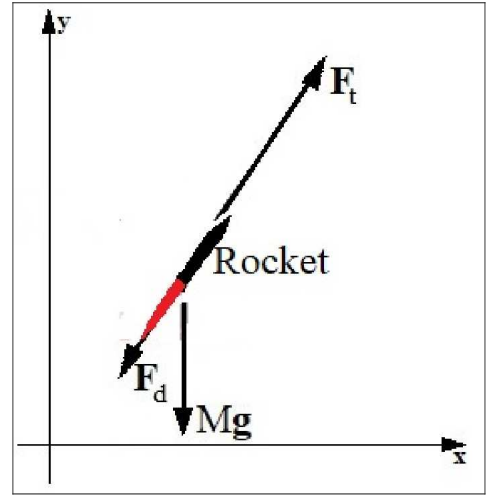}
\captionof{figure}{Forces acting on a firing rocket \label{F03}}
\end{wrapfigure}

Three forces acting on a rocket during flight are presented in Figure \ref{F03}.  The gravitational force is $M(t) \mathbf{g}$ pointed in $-\hat{y}$ direction.  Notice, that while the rocket fuel is burning, the rocket mass is decreasing.  The air resistance or drag force $\mathbf{F}_d$ is acting in the opposite direction of velocity.  The magnitude of $F_d$ is
\begin{linenomath} \be
\label{2.01}
\begin{split}
F_d&=C_d(\mathcal{M})\ \ \frac{\rho v^2 A}{2}\\
&=C_d(\mathcal{M})
\ \ \mathcal{M}^2 \ \ \frac{\rho v_s^2 A}{2},
\end{split}
\ee \end{linenomath}
where $C_d(\mathcal{M})$ is the Mach--number dependent drag coefficient, $v$ is the velocity, $v_s$ is the speed of sound, $\mathcal{M}$ is the Mach number, i.e. $\mathcal{M}=v/v_s$, $\rho$ is the air density, and $A$ is the base area of the rocket.  Let $M$ be the rocket mass.  Then thrust force $\mathbf{F}_t$  acting in the direction of velocity is given by
\begin{linenomath} \be
\label{2.02}
F_t=\dot{M} v_e,
\ee \end{linenomath}
where $\dot{M}$ is the fuel burning rate, and $v_e$ is the equivalent exhaust velocity.

The exhaust velocity increases as the rocket rises out of dense atmospheric layers.  For the DLSRR launched at a high altitude, $v_e$ barely changes.  For some propellant grains, the fuel burning rate changes during the burn time.  In our case we assume an almost steady burning rate
\begin{linenomath} \be
\label{2.03}
\dot{M}=\frac{M\ f_p}{t_b} \qquad \text{for time} \ \ 0 \le t \le t_b,
\ee \end{linenomath}
where $t_b$ is the burn time and $f_p$ is the propellant mass fraction given by
\begin{linenomath} \be
\label{2.04}
f_p=\frac{\text{Propellant mass}}{
\left\{
\begin{split}
&\text{Combined mass of propellant,}\\
&\text{rocket, and payload}
\end{split}
\right\}
}.
\ee \end{linenomath}
Since we know the forces acting on the rocket, we calculate its trajectory via the MatLab programs \textbf{FirstStage.m} and \textbf{Rocket.m} , which are described in the Appendix.

The change in velocity produced by the rocket engine is
\begin{linenomath} \be
\label{2.05}
v_r=\int_0^{t_b} \frac{\mathbf{F}_t(t)}{M(t)}\ dt,
\ee \end{linenomath}
where $M(t)$ is the rocket mass at time $t$.
According to \emph{Tsialkovski Rocket Equation} \cite{Tsialkovski},
\begin{linenomath} \be
\label{2.06}
v_r=-\overline{v}_e\ ln \big(1-f_p \big),
\ee \end{linenomath}
where $\overline{v}_e$ is the average exhaust velocity.  In our case,
$\overline{v}_e \approx 2,100\ m/s$.  For expensive rockets with flame temperatures in excess of 2,800 $^oC$, $\overline{v}_e \approx 2,600\ m/s$.

The drag loss is
\begin{linenomath} \be
\label{2.07}
v_d=\int_0^{t_0} \frac{F_d(t)}{M(t)}\ dt,
\ee \end{linenomath}
where $M(t)$ is the rocket mass at time $t$, $t_0$ is the time it takes the rocket to reach the apogee, and $F_d(t)$ is the drag force at the time moment $t$.

At this point we define the effective loss of rocket velocity due to gravity.  It is called gravitational loss and denoted $v_g$.  We define this loss in terms of the rocket's speed and altitude at the apogee.  First, assume that the rocket is given its impulse $v_r$ instantaneously, and the drag loss is negligible.  Such assumption is an abstract limit for a rocket, but it may be reality for a projectile fired at high altitude.  Then the total kinetic and potential energy per unit rocket (projectile) mass at the beginning of trajectory is
\begin{linenomath} \be
\label{2.08}
\mathcal{E}=g h_0+\frac{\big(v_0+v_r\big)^2}{2},
\ee \end{linenomath}
where $h_0$ is the launch altitude, $v_0$ is the launch velocity, and $v_r$ is the velocity gain due to the action of the rocket engine.  Second, we incorporate the aerodynamic drag loss into (\ref{2.08}) to obtain
\begin{linenomath} \be
\label{2.09}
\mathcal{E}=g h_0+\frac{\big(v_0+v_r-v_d\big)^2}{2},
\ee \end{linenomath}
where $v_d$ is the drag loss.  Third, we incorporate the gravitational loss into (\ref{2.09}) to obtain
\begin{linenomath} \be
\label{2.10}
\mathcal{E}=g h_0+\frac{\big(v_0+v_r-v_d-v_g\big)^2}{2},
\ee \end{linenomath}
where $v_g$ is the gravitational loss.  In the above expressions, $v_0$, $v_r$, $v_d$, and $h_0$ are known, while both $\mathcal{E}$ and $v_g$ are unknown.  In order to calculate $v_g$, we calculate $\mathcal{E}$ using the parameters of rocket motion at the apogee.  The maximum ascent of the rocket, or the height the rocket attains at apogee is denoted $h_{_A}$.  The rocket speed at the apogee is $v_{_A}$.  The vector of this speed is in the forward direction.  The total kinetic and potential energy per unit rocket mass at the apogee is
\begin{linenomath} \be
\label{2.11}
\mathcal{E}=g h_{_A}+\frac{v_{_A}^2}{2},
\ee \end{linenomath}
which is the same as the rocket energy earlier in the path given in (\ref{2.10}).  Equating (\ref{2.10}) and (\ref{2.11}), we obtain
\begin{linenomath} \be
\label{2.12}
v_g=\big(v_0+v_r\big)-v_d-\sqrt{2 g \big(h_{_A}-h_0 \big)+v_{_A}^2}.
\ee \end{linenomath}

\section{DLSRR Engines}
\subsection{Solid Propellant Rockets}
In the present and the two following sections we consider a solid propellant rocket.  It is schematically presented in Figure \ref{F04} below.
\begin{center}
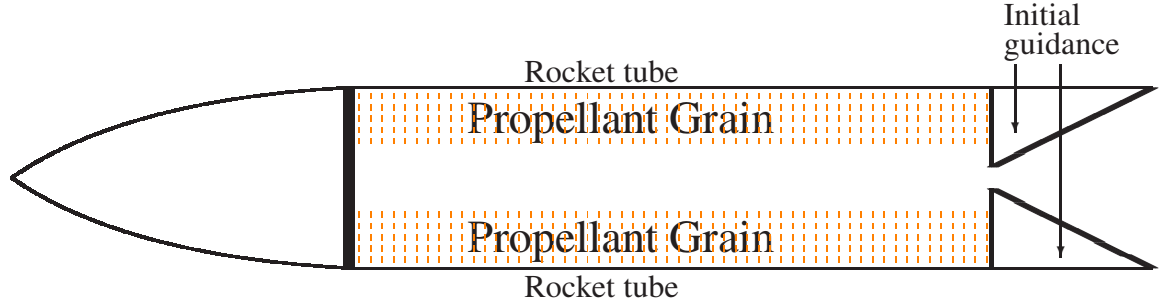

\setlength{\unitlength}{1.5mm}
\begin{picture}(100, 25)
\linethickness{0.3mm}

\put(30,2){\line(1,0){70}}
\put(30,18){\line(1,0){70}}
\multiput(30,2)(-.2,0){5}{\line(0,1){16}}
\put(30,2){\line(0,1){16}}
\qbezier(0,10)(10,3)(30,2)
\qbezier(0,10)(10,17)(30,18)

\put(45,18.5){Rocket tube}
\put(45,-0.5){Rocket tube}

\multiput(86.5,11)(-.1,0){10}{\line(2,1){14}}
\multiput(86.5,9)(-.1,0){10}{\line(2,-1){14}}

\multiput(86,11)(-.1,0){3}{\line(0,1){7}}
\multiput(86,9)(-.1,0){3}{\line(0,-1){7}}

\linethickness{0.2mm}
\put(88,20){\vector(0,-1){6}}
\put(92,20){\vector(0,-1){17}}
\put(87,23.5){Initial}
\put(87,21){guidance}

\put(40,14){\Large Propellant Grain}
\put(40,3.5){\Large Propellant Grain}

\linethickness{0.1mm}
\multiput(30.5,13)(0,1){5}{\color[rgb]{1,.5,0}{\multiput(0,0)(1,0){56}{\line(0,1){0.5}}}}
\multiput(30.5,7)(0,-1){5}{\color[rgb]{1,.5,0}{\multiput(0,0)(1,0){56}{\line(0,-1){0.5}}}}

\end{picture}
\captionof{figure}{Solid propellant rocket \label{F04}}
\end{center}

The DLSRR should consist of a rocket with one or more guided warheads.  These rockets would be much less expensive than conventional rockets with similar range and payload.  First, the fact that the DLSRR starts at an altitude of over 12 $km$ to 20 $km$ with a velocity 500 $m/s$ to 700 $m/s$ would reduce $\triangle v$ needed for the rocket to achieve a given range by at least 750 $m/s$.  The reduction in needed $\triangle v$ is greater than the initial velocity due to the fact that the rocket starts out at high altitude.  First, it decreases the drag loss.  Second, initial elevation itself reduces $\triangle v$ needed to achieve a given range. Second, as we show in the following sections, a rocket starting at high altitude with high initial velocity does not need to experience high firing acceleration.  While most artillery rockets burn their fuel within 2-3 seconds, the DLSRR will be able to achieve good result even if it takes 40-60 seconds to burn its fuel.  A solid-propellant rocket with slow burning fuel is much less expensive to manufacture.  Third, slow-burning fuels have much lower flame temperature \cite{pbrt}, which greatly reduces the cost of exhaust and initial guidance.
There is a general pattern that propellants with higher combustion temperatures have higher burning rates than propellants with lower combustion temperatures at the same pressure.  For example, ammonium nitrate propellants have low burning rates.  Even though exceptions exist, they are rare.  This phenomenon has a physical explanation.  Lower temperature flames send lower heat flux back to the propellant.  Lower heat flux causes the propellant to pyrolyse at lower rate.  Thus, the overall regression and burning rate of the propellant is lower.

The initial guidance of DLSRR should be simple.  It must make sure that DLSSR does not sway off the course by more than a kilometer.  Probably inertial guidance with fins within the nozzle can accomplish the task.

\subsection{Propellant Composition}
Most modern artillery rockets as well as space rocket boosters use a solid propellant containing 70\% ammonium perchlorate (AP), 15\% aluminum powder, and 15\% HTPB binder \cite[p.479]{RPE}.  Such propellant is expensive to manufacture due to the fact that AP is highly explosive.  Other propellants contain high explosives and highly nitrated nitrocellulose.  All of them are expensive and dangerous to handle.  All the aforementioned propellants have flame temperatures of 2,500 $^oC$ to 3,500 $^oC$.  Such exhaust temperatures require expensive nozzles.

DLSRR have a choice of fuels burning at lower temperature and rate.  Some formulations contain ammonium nitrate (AN).  The chemical formula for ammonium nitrate is
$NH_4NO_3$.  Its heat of formation is 367 $kJ/mol$.  Ammonium Nitrate density is 1.725 $g/cm^3$ \cite[p. 256]{PSAN}.
Ammonium Nitrate undergoes phase transitions at -18 $^oC$ and +32 $^oC$.  These phase transitions are accompanied by volume change of about 4\%, thus they must be avoided in order to avoid fuel grain destruction \cite[p. 263]{PSAN}.  Ammonium Nitrate can be phase-stabilized by addition of 10\% potassium nitrate or 2\% potassium fluoride \cite[p. 267-268]{PSAN}.  

Other formulations have mildly nitrated nitrocellulose.  The chemical formula for Nitrocellulose is
$C_6 H_{10-x} O_5 \big(N O_2 \big)_x$.  Its atomic mass is $162+45 x$.  Its binding energy is \\ $\big( 961.5-103.6 \cdot x \big) kJ/mol$.
The number $0<x<3$ determines the oxidizer content of nitrocellulose.
Nitroglycerin (NG) is used in double base propellants with nitrocellulose \cite{Double}.  Nitroglycerin formula is $H_5 C_3 N_3 O_9$.
Its atomic mass is 227 amu.
Its heat of formation is 380 $kJ/mol$.
The propellants we are interested in should have flame temperatures slightly below 1,500 $^oC$.

In Table \ref{T03} below, we list several propellants.  In the first column, we list propellant composition.  In the second column, we list the flame temperature.  In the third column, we list exhaust velocity into vacuum given an initial pressure of 40 $atm$ and expansion of 15.  Both temperature and the exhaust velocity are calculated using the program called Rocket Propulsion Analsis \cite{RPA}, which is available online.  The forth column lists the burning rate at 40 $atm$.  The fifth column lists power coefficient.  For almost all propellants, the burning rate is approximated by
\begin{linenomath} \be
\label{3.01}
r_b\big(P \big)=r_b \big( P_{_0} \big) \left( \frac{P}{P_{_0}} \right)^{n},
\ee \end{linenomath}
where $r_b$ is the burning rate, $P$ is pressure, $P_{_0}$ is the reference pressure, and $n$ is the power coefficient.
\begin{center}
\begin{tabular}{|l|l|l|l|l|l|l|l|l|l|l|l|}
  \hline
  Propellant  & Temperature & $v_e$ at 40 $atm$,   & Burning & $n$ coefficient \\
  Composition & at 40 $atm$ &  Exp 15              & rate    & at 40 $atm$     \\
              &             & Vacuum               &         &                 \\
  \hline
  20\% Binder, 72\% AN, 8\% MgAl  \cite{ANMg2}& 1,280 $^oC$ & 2,140 $m/s$ & 2.0 $mm/s$ &  0.5  \\
  20\% Binder, 68\% AN, 12\% MgAl             & 1,450 $^oC$ & 2,240 $m/s$ & NA         &   NA  \\
  20\% Binder, 64\% AN, 16\% MgAl \cite{ANMg2}& 1,700 $^oC$ & 2,320 $m/s$ & 3.0 $mm/s$ &  0.5  \\
  25\% Binder, 60\% AN, 15\% MgAl \cite{ANMg2}& 1,410 $^oC$ & 2,170 $m/s$ & 1.6 $mm/s$ &  0.7  \\
  \hline
  17.8\% HTPB, 64\% AN,   & 1,660 $^oC$ & 2,270 $m/s$ & 2.0 $mm/s$ & 0.05 \\
  14.6\% Mg, 3.6\% AC  \cite{ANMg1}  &             &             &            &      \\
  \hline
  Nitrocellulose  9\% N                 & 1,060 $^oC$ & 1,990 $m/s$ & NA & NA \\
  Nitrocellulose 10\% N                 & 1,350 $^oC$ & 2,090 $m/s$ & NA & NA \\
  Nitrocellulose 11\% N                 & 1,680 $^oC$ & 2,200 $m/s$ & NA & NA \\
  Nitrocellulose 12\% N                 & 2,030 $^oC$ & 2,290 $m/s$ & NA & NA \\
  Nitrocellulose 13\% N                 & 2,380 $^oC$ & 2,420 $m/s$ & NA & NA \\
  Nitrocellulose 14\% N                 & 2,690 $^oC$ & 2,530 $m/s$ & NA & NA \\
  \hline
  90\% Nitrocellulose  7.5\% N, 10\% Mg & 1,290 $^oC$ & 2,080 $m/s$ & NA & NA \\
  90\% Nitrocellulose  8\%   N, 10\% Mg & 1,410 $^oC$ & 2,130 $m/s$ & NA & NA \\
  90\% Nitrocellulose  8.5\% N, 10\% Mg & 1,540 $^oC$ & 2,200 $m/s$ & NA & NA \\
  \hline
  70\% AP, 15\% Al, 15\% HTPB & 2,880 $^oC$ & 2,610 $m/s$ & 6.5 $mm/s$ & 0.35 \\
  Standard Propellant \cite{ApAl1}       & & & & \\
  \hline
  56\% NG, 44\% Cellulose  & 1,340 $^oC$             & 2,090 $m/s$               & NA         & NA \\
  59\% NG, 41\% Cellulose  & 1,510 $^oC$             & 2,150 $m/s$               & NA         & NA \\
  66\% NG, 34\% Cellulose \cite[p. 4-5]{Double2}     & 1,950 $^oC$ & 2,300 $m/s$ & 4.0 $mm/s$ & 0.7 \\
  77\% NG, 23\% Cellulose \cite[p. 4-5]{Double2}     & 2,560 $^oC$ & 2,530 $m/s$ & 7.0 $mm/s$ & 0.7 \\
  82.5\% NG, 17.5\% Cellulose \cite[p. 4-5]{Double2} & 2,790 $^oC$ & 2,630 $m/s$ & 10  $mm/s$ & 0.7 \\
  \hline
\end{tabular}
\captionof{table}{Performance of solid propellants (NA -- not available) \label{T03}}
\end{center}


Propellants using ammonium nitrate oxidizer may experience burn rate instability for low and moderate pressures.  More research is needed in order to determine feasibility and cost of using such propellants.  Ammonium Dinitramide and Hydrazinium Nitroformate have been proposed as solid propellant oxidizers.  Propellant formulations using these oxidizers have excellent performance.  The problem with these oxidizers is their cost of hundreds of dollars per kilogram.

Finding a propellant optimal in terms of both cost and properties remains an open problem.  For now we assume, that our propellant has flame temperature 1,450 $^oC$, burning rate of 2.0 $mm/s$, and exhaust velocity of 2,100 $m/s$.  Extrapolation of the data in \cite[p.151]{PropExp} also predicts the aforementioned combination of burning rate and exhaust velocity.
Using the densities of energetic materials in \cite[p.37]{PropExp}, the density of our propellant should be 1.6 $g/cm^3$.

\subsection{Grain Shape}
The grain shape is the shape of the fuel within the rocket tube.  Different rockets use a wide variety of grain shapes illustrated in Figure \ref{F05} below:
\begin{center}
\includegraphics[width=12cm,height=7cm]{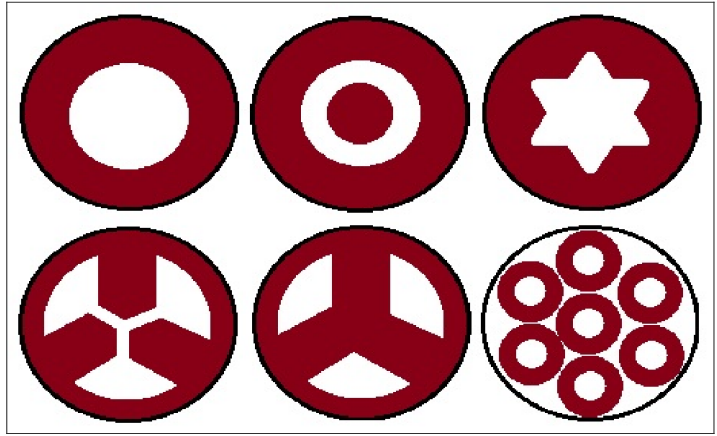}
\captionof{figure}{Propellant grain shapes \label{F05}}
\end{center}

The shape may change over the length of the rocket tube.
We mention that different types of grain shapes provide different thrust curves.  In the present paper, we focus on grain shape providing constant thrust.
Propellant loading is the fraction of rocket tube volume occupied by the propellant.  The best propellant grain shape for the DLSRR satisfies three requirements.  First, the burning propellant surface area should experience minimal change as the propellant is burning.  This would ensure steady thrust.  Second, the propellant loading should be as high as possible.  This would maximise fuel mass ratio within the rocket.  Third, the design should be as simple as possible.  This would minimize cost.

A diagram of a rocket motor cross-section is below:
\begin{center}
\includegraphics[width=12cm,height=7cm]{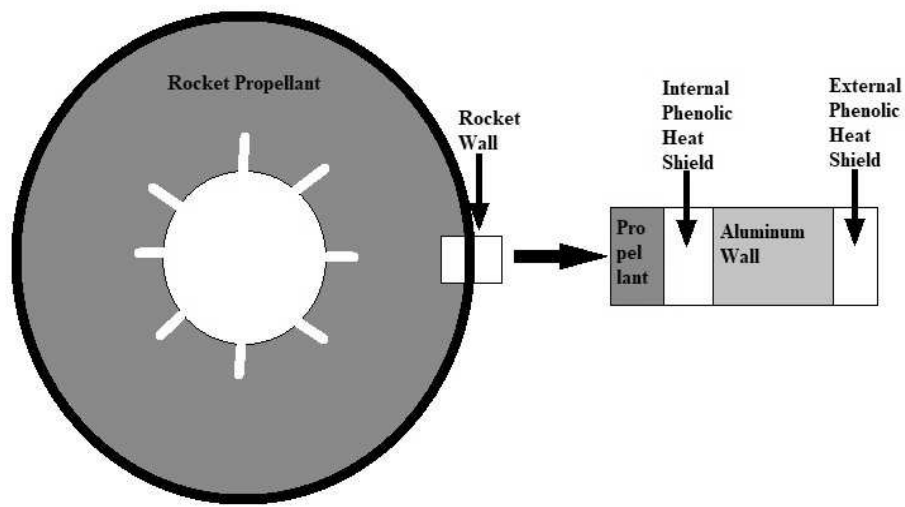}
\captionof{figure}{Rocket motor cross section \label{F06}}
\end{center}
Aluminum 6061 T6 is one of the best candidates for the rocket wall.
Aluminum 6061 T6 pipe of diameter up to 25 $cm$ and wall thickness 1 $cm$ is sold for  $\$14.30$ per $kg$ pipe weight \cite{OnlineMetals}.  It has density of 2.7 $g/cm^3$ and tensile yield strength of 2,720 $atm$ \cite{Al6061T6v2}.  The mass of solid propellant rocket wall for a given volume and pressure is inversely proportional to the wall's specific strength, which is yield strength divided by density.   Specific strength has units of
\begin{linenomath} \be
\label{3.02}
\frac{N/m^2}{kg/m^3}=\frac{N \cdot m}{kg}=\frac{J}{kg}.
\ee \end{linenomath}
Aluminum 6061 T6 has a specific strength of $1.15 \cdot 10^5\ J/kg$.
Duralumin and titanium have specific strength about twice as high as Aluminum 6061 T6, but they are tens of times more expensive than Aluminum 6061 T6.  Composite materials with specific strength of $1.7 \cdot 10^5\ J/kg$ and price of \$60 per $kg$ in 2009 are available.  These materials are widely used in compressed natural gas tanks \cite{CNG}.  Currently, these tanks are for sale at about \$60 per $kg$ wall material.

\section{Rocket And Warhead Performance}
\subsection{Performance of 30 $cm$ Drone Launched Short Range Rockets (DLSRR)}
The rocket is shown below:

\begin{center}
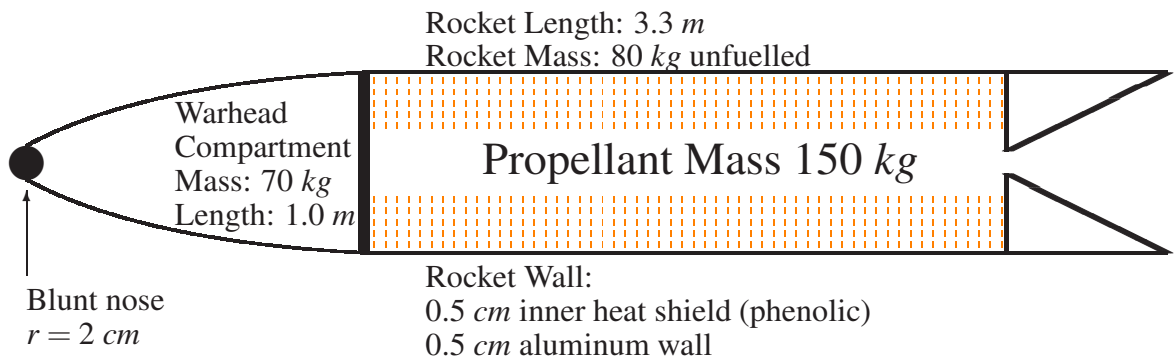

\setlength{\unitlength}{1.5mm}
\begin{picture}(100, 35)
\linethickness{0.3mm}

\put(30,12){\line(1,0){70}}
\put(30,28){\line(1,0){70}}
\multiput(30,12)(-.2,0){5}{\line(0,1){16}}
\put(30,12){\line(0,1){16}}
\qbezier(0,18.5)(10,13)(30,12)
\qbezier(0,21.5)(10,27)(30,28)

\put(0,20){\circle*{3}}
\linethickness{0.1mm}
\put(0,10){\vector(0,1){7.5}}
\linethickness{0.3mm}
\put(0,7){Blunt nose}
\put(0,4){$r=2\ cm$}

\put(13,23.5){Warhead}
\put(13,20.5){Compartment}
\put(13,17.5){Mass: 70 $kg$}
\put(13,14.5){Length: 1.0 $m$}

\put(35,31.5){Rocket Length: 3.3 $m$}
\put(35,28.5){Rocket Mass:   80 $kg$ unfuelled}

\put(35,9.0){Rocket Wall:}
\put(35,6.0){0.5 $cm$ inner heat shield (phenolic)  }
\put(35,3.0){0.5 $cm$ aluminum wall }

\multiput(86.5,21)(-.1,0){10}{\line(2,1){14}}
\multiput(86.5,19)(-.1,0){10}{\line(2,-1){14}}

\multiput(86,21)(-.1,0){3}{\line(0,1){7}}
\multiput(86,19)(-.1,0){3}{\line(0,-1){7}}

\put(40,19){\Large Propellant Mass 150 $kg$}

\linethickness{0.1mm}
\multiput(30.5,23)(0,1){5}{\color[rgb]{1,.5,0}{\multiput(0,0)(1,0){56}{\line(0,1){0.5}}}}
\multiput(30.5,17)(0,-1){5}{\color[rgb]{1,.5,0}{\multiput(0,0)(1,0){56}{\line(0,-1){0.5}}}}

\linethickness{0.3mm}
\end{picture}
\captionof{figure}{Drone launched short range rocket \label{F07}}
\end{center}

The rocket has diameter of 30 $cm$, total mass of 300 $kg$, total length of 4.3 $m$ of which 1.0 $m$ is the ogive warhead compartment.  The propellant mass fraction is $f_p=0.50$.   We took typical values for a 30 $cm$ rocket, that can be found in the literature.  The exhaust velocity is  $v_e = 2,100\ m/s$.

Using RASAero software \cite{rasaero} we have calculated the drag coefficient for aforementioned artillery rocket at different Mach numbers and presented in Table \ref{T04} below.  The RASAero software takes in rocket radius, rocket cylinder length, rocket nose length, and rocket nose shape.  The software calculates the drag coefficient.
\begin{center}
\begin{tabular}{|c|c|c|c|c|c|c|c|c|c|c|c|c|c|c|c|c|c|c|}
  \hline
  $\mathcal{M}$   &.25   &.50   &.75   & 1.00  &1.25  & 1.50  & 1.75 & 2.00  & 2.25 & 2.50  & 2.75 & 3.00 \\
  \hline
  $C_d$ & .265 & .260 & .261 & .380  &.432  & .404  & .381 & .352  & .328 & .308  & .289 & .273 \\
  \hline
  \hline
  \hline
    $\mathcal{M}$ &3.25  &3.50  &3.75  & 4.00  &4.25  & 4.50  & 4.75 & 5.00  & 5.25 & 5.50  & 5.75 & 6.00 \\
  \hline
  $C_d$ & .259 & .246 & .235 & .225  & .216 & .208  & .200 & .193  & .187 & .181  & .175 & .168 \\
  \hline
  \hline
  \hline
    $\mathcal{M}$ &6.25  &6.50  &6.75  & 7.00  &7.25  & 7.50  & 7.75 & 8.00  & 8.25 & 8.50  & 8.75 & 9.00 \\
  \hline
  $C_d$ & .163 & .159 & .154 & .150  & .147 & .144  & .141 & .139  & .137 & .135  & .133 & .131 \\
  \hline
\end{tabular}
\captionof{table}{Drag coefficient for 30 $cm$ rocket \label{T04}}
\end{center}

The warhead compartment consists of a 10 $kg$ container which ejects up to 60 $kg$ of projectiles at the apogee.  The projectiles are guided.  They are made of high density material like tungsten or depleted uranium.

Performance of DLSRR for several release altitudes, release velocities, and firing times is present in Table \ref{T05} below.  Three combinations of release altitude and velocity are selected as representative examples.  Other combinations are possible. We have presented several grain shapes and propellant compositions.  Any combination can be chosen, thus different firing times are possible.  For each set of release altitude, release velocity, and propellant burning time, we have calculated the ranges for different firing angles and selected the one giving the longest range.  The first row is the engine firing time.  The second row is the firing angle producing the greatest range.  Rows 3-7 are self-explanatory.  Row 8 is the expected range of a warhead released at the apogee. All results have been calculated by Rocket.m program described in Appendix.
\begin{center}
\begin{tabular}{|l|r|r|r|r|r|r|r|r|r|r|r|r|r|r|}
  \hline
  \multicolumn{6}{|c|}{\textbf{Release altitude: 12 $km$.  Release velocity: 500 $m/s$.}} \\
  \hline
  Firing time, $s$            &   30 &   40 &   50 &   60 &   80 \\
  Optimal firing angle, $^o$  &   54 &   57 &   61 &   64 &   68 \\
  Flight time, $s$            &  282 &  288 &  289 &  289 &  277 \\
  Drag loss, $m/s$            &   98 &   84 &   76 &   69 &   62 \\
  Gravity loss, $m/s$         &   93 &  131 &  169 &  209 &  285 \\
  Rocket Velocity Gain, $m/s$ & 1457 & 1457 & 1457 & 1456 & 1459 \\
  Flight apogee, $km$         &   93 &   94 &   92 &   89 &   77 \\
  Range, $km$                 &  327 &  317 &  304 &  291 &  263 \\
  \hline
\end{tabular}

\begin{tabular}{|l|r|r|r|r|r|r|r|r|r|r|r|r|r|r|}
  \hline
  \multicolumn{6}{|c|}{\textbf{Release altitude: 16 $km$.  Release velocity: 600 $m/s$.}} \\
  \hline
  Firing time, $s$            &   30 &    40 &   50 &   60 &   80 \\
  Optimal firing angle, $^o$  &   54 &    57 &   59 &   60 &   65 \\
  Flight time, $s$            &  311 &   314 &  311 &  303 &  300 \\
  Drag loss, $m/s$            &   51 &    45 &   42 &   40 &   35 \\
  Gravity loss, $m/s$         &   90 &   124 &  158 &  188 &  263 \\
  Rocket Velocity Gain, $m/s$ & 1457 &  1457 & 1457 & 1456 & 1456 \\
  Flight apogee, $km$         &  116 &   115 &  110 &  102 &   95 \\
  Range, $km$                 &  386 &   373 &  361 &  349 &  318 \\
  \hline
\end{tabular}

\begin{tabular}{|l|r|r|r|r|r|r|r|r|r|r|r|r|r|r|}
  \hline
  \multicolumn{6}{|c|}{\textbf{Release altitude: 20 $km$.  Release velocity: 700 $m/s$.}} \\
  \hline
  Firing time, $s$            &   30 &    40 &   50 &   60 &   80 \\
  Optimal firing angle, $^o$  &   53 &    54 &   56 &   58 &   62 \\
  Flight time, $s$            &  320 &   325 &  323 &  322 &  318 \\
  Drag loss, $m/s$            &   29 &    26 &   24 &   22 &   20 \\
  Gravity loss, $m/s$         &   82 &   113 &  144 &  176 &  242 \\
  Rocket Velocity Gain, $m/s$ & 1457 &  1457 & 1457 & 1456 & 1459 \\
  Flight apogee, $km$         &  125 &   126 &  122 &  118 &  111 \\
  Range, $km$                 &  444 &   431 &  417 &  403 &  374 \\
  \hline
\end{tabular}
\captionof{table}{Performance of 30 $cm$ rocket \label{T05}}
\end{center}
For some simulations there is a small roundoff error of up to $2\ m/s$.  There may be a very slight variance in $\triangle v$ due to the fact that exhaust velocity depends on ambient air pressure.
DLSRR trajectories are plotted in Figure \ref{FN1} below:
\begin{flushleft}
\includegraphics[width=15cm]{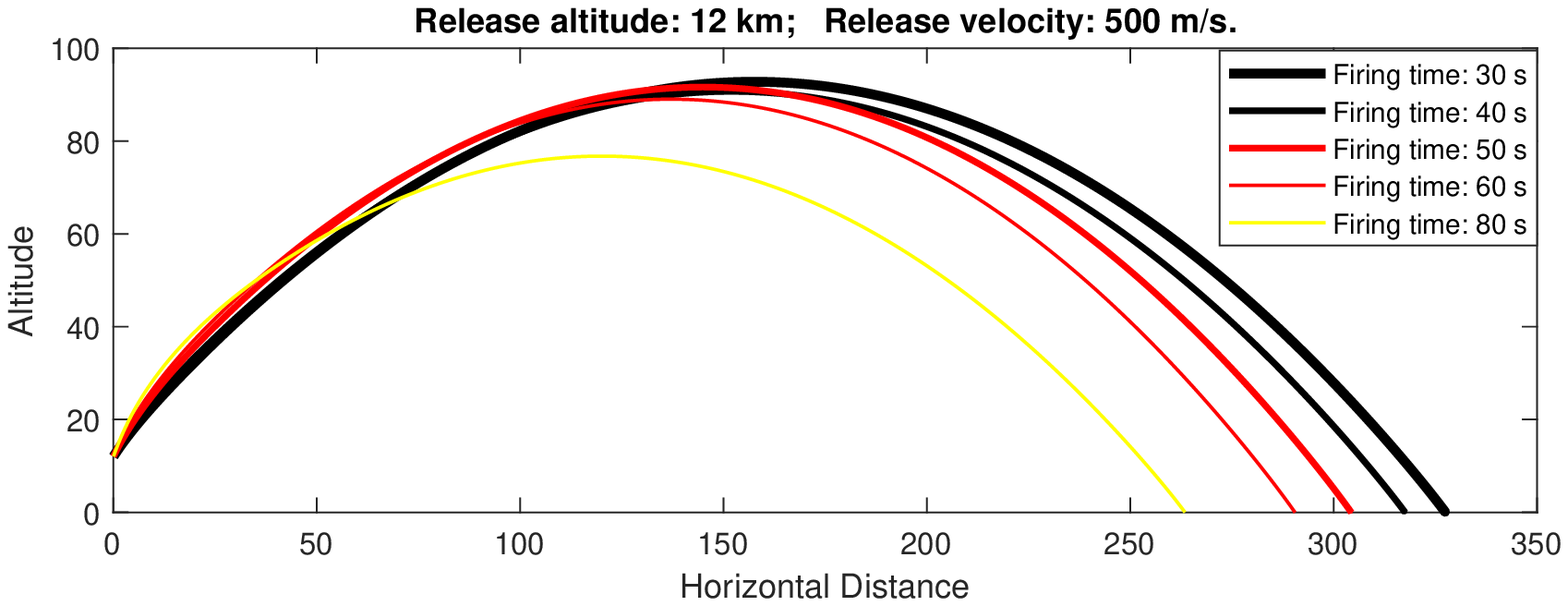}
\includegraphics[width=15cm]{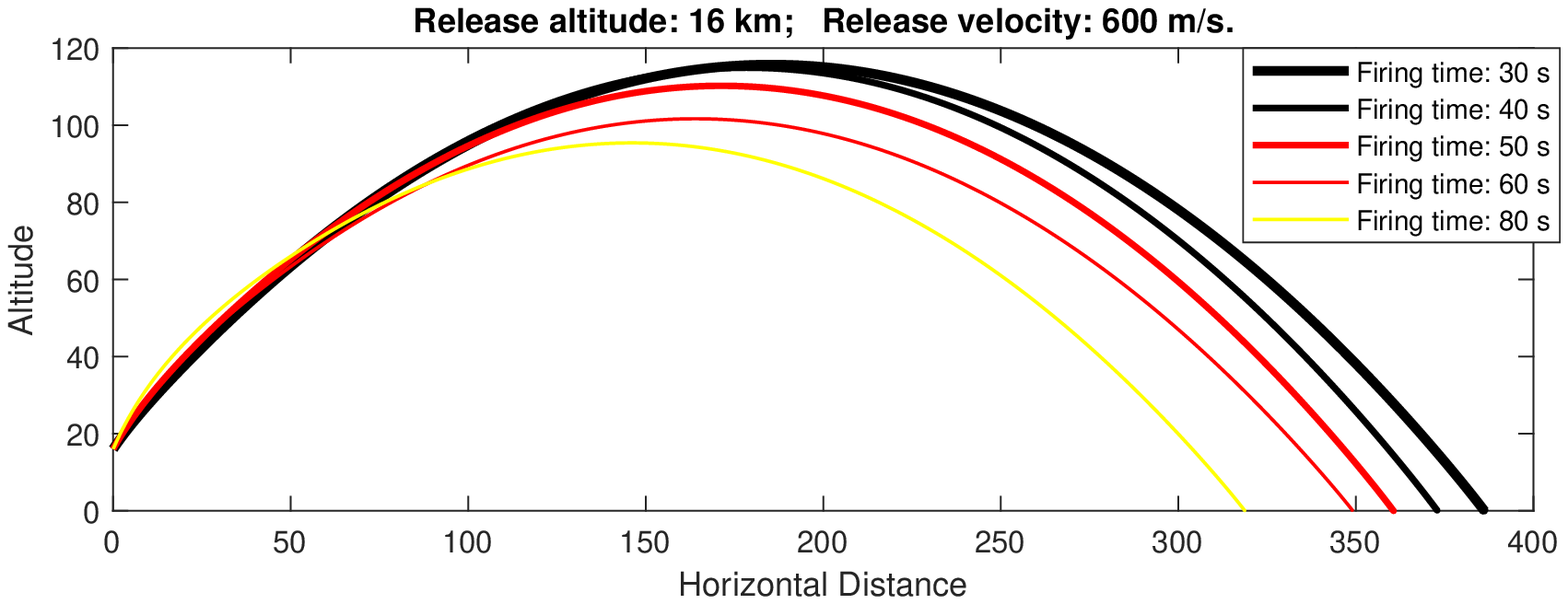}
\includegraphics[width=15cm]{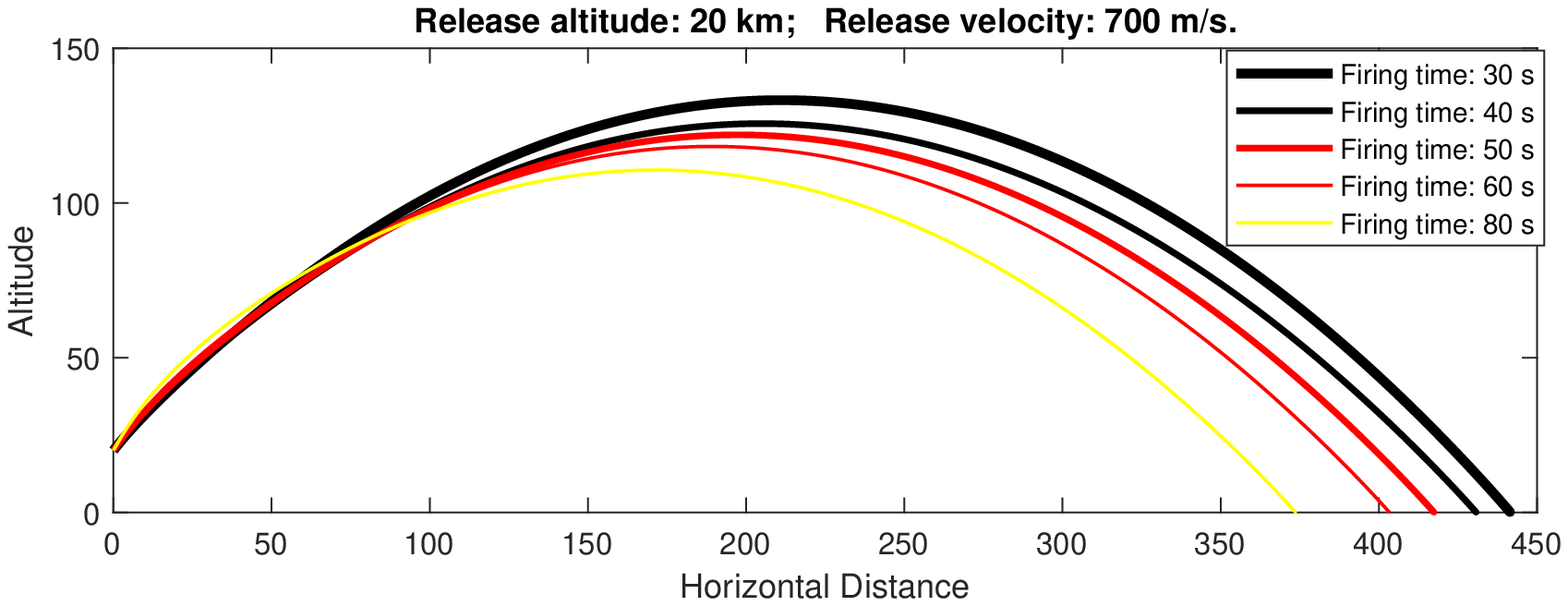}
\captionof{figure}{DLSRR trajectories \label{FN1}}
\end{flushleft}

\subsection{Impact Velocity Calculation}
\subsubsection{Hypervelocity Projectile}

In this subsection we calculate the impact velocity for the warhead consisting of the Hypervelocity Projectile (HPV).  HPV is described in Subsection 1.2 and presented in Fig. \ref{F02}.   Recall, that HPV's weight is 11.4 $kg$, length 64.2 $cm$, and caliber 7.83 $cm$.  A 70 $kg$ warhead compartment can carry up to 5 such projectiles.

Using RASAero software \cite{rasaero} we have calculated the drag coefficient for the HPV at different Mach numbers and presented in Table \ref{T06} below.  As we mentioned earlier, the RASAero software takes in rocket radius, rocket cylinder length, rocket nose length, and rocket nose shape.  The software calculates the drag coefficient.
\begin{center}
\begin{tabular}{|c|c|c|c|c|c|c|c|c|c|c|c|c|c|c|c|c|c|c|}
  \hline
  $\mathcal{M}$   &.25   &.50   &.75   & 1.00  &1.25  & 1.50  & 1.75 & 2.00  & 2.25 & 2.50  & 2.75 \\
  \hline
  $C_d$ & .208 &.203 & .202 & .289 & .336 & .325 & .292 & .265 & .243 & .225 & .208 \\
  \hline
\end{tabular}\\
\begin{tabular}{|c|c|c|c|c|c|c|c|c|c|c|c|c|c|c|c|c|c|c|}
  \hline
  $\mathcal{M}$    & 3.00 & 3.25 & 3.50 & 3.75 & 4.00 &4.25  &4.50  &4.75  & 5.00  &5.25 & 5.50  \\
  \hline
  $C_d$ & .194& .181 & .171 & .161& .152 &.145  &.139 &.133  & .127 &.122 & .117\\
  \hline
\end{tabular}\\
\begin{tabular}{|c|c|c|c|c|c|c|c|c|c|c|c|c|c|c|c|c|c|c|}
  \hline
  $\mathcal{M}$      & 5.75 & 6.00  & 6.25 & 6.50  & 6.75 & 7.00 & 7.25 & 7.50 & 7.75 & 8.00 \\
  \hline
  $C_d$    & .112 &.106  & .103  & .099&.095  &.091 &.089 &.087 &.085 & .083 \\
  \hline
\end{tabular}
\captionof{table}{Drag coefficient for HPV projectile \label{T06}}
\end{center}

In Rocket.m simulations described in Subsection 4.1, we have calculated DLSRR apogee altitude, DLSRR apogee velocity, and DLSRR apogee horizontal distance.  These calculations were performed for release altitudes of 12 $km$, 16 $km$, and 20 $km$.  Corresponding release velocities were 500 $m/s$, 600 $m/s$, and 700 $m/s$.  We input the results of Rocket.m simulations and the drag coefficient tabulated in Table \ref{T06} above into Impact.m.  The program Impact.m calculates the projectiles' ranges, impact velocities, and descent drag loss.
The results of calculations are presented in Table \ref{T07} below.
The first row is the initial velocity.
The second row is the firing altitude.
The third row is the rocket firing time.
The fourth row is the apogee altitude.
The fifth row is the apogee velocity.
The sixth row is the warhead horizontal range from the firing point.
The seventh row is the impact velocity.
The eighth row is the descent drag loss.

\begin{center}
\begin{tabular}{|l|r|r|r||r|r|r||r|r|r|r|r|r|r|r|}
  \hline
  DLSRR $v_0$, $m/s$   &  500  &   600 &   700  &  500  &   600 &   700  &  500  &   600 &  700  \\
  DLSRR $h_0$, $km$    &   12  &    16 &    20  &   12  &    16 &    20  &   12  &    16 &   20  \\
  DLSRR $t_f$, $s$     &   30  &    30 &    30  &   50  &    50 &    50  &   80  &    80 &   80  \\
  \hline
  \hline
  Apogee $h$, $km$     &    93 &   116 &    133 &    92 &   110 &   122  &    77 &    95 &   111 \\
  Apogee $v$, $m/s$    & 1,239 & 1,310 &  1,399 & 1,170 & 1,265 & 1,398  & 1,151 & 1,241 & 1,349 \\
  \hline
  \hline
  Range, $km$          &   328 &   386 &    442 &   305 &   360 &   418  &   264 &   319 &   374 \\
  Impact $v$, $m/s$    & 1,252 & 1,422 &  1,567 & 1,215 & 1,372 & 1,501  & 1,110 & 1,269 & 1,409 \\
  $\triangle v_{_{\text{loss}}}$, $m/s$  &   580 &   574 &    571 &   565 &   568 &   584  &   573 &   579 &   589 \\
  \hline
\end{tabular}
\captionof{table}{HPV warhead performance \label{T07}}
\end{center}

As we see from the above table, the HPV has very high aerodynamic drag loss.  Even though minimization of drag loss has been an important goal in the design of HPV, the main design goal for HPV was an ability to survive the acceleration of a cannon launch.  Rocket launched projectiles do not require the ability to survive extreme acceleration, hence HPV design may be suboptimal for rocket-launched projectiles.

\subsubsection{European Hypersonic Projectile}
The Europrojectile has not been built yet and exists only as a concept.  Based on paper \cite{EurRail}, we present a sketch below.

\begin{center}
\includegraphics[width=15cm]{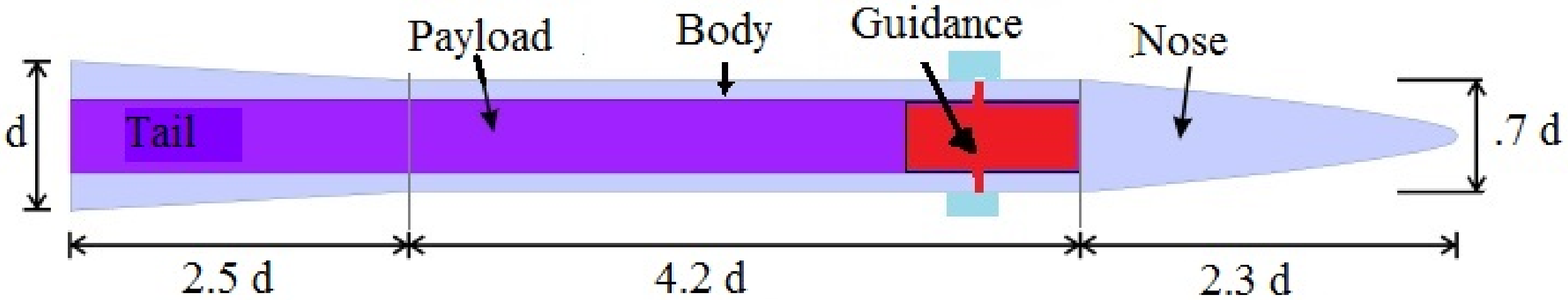}
\captionof{figure}{Europrojectile \label{F08}}
\vskip0.5cm
\end{center}

In the above figure, $d$ is the base diameter.  We will consider several diameters.  The length of the projectile is $9\ d$.  The tail of the projectile is a conical frustum with height $2.5\ d$ and radii $0.5\ d$ and $0.35\ d$.  The volume of the tail is
\begin{linenomath} \be
\label{4.01}
\frac{\pi}{3} H\left(R_1^2+ R_1 R_2+ R_2^2 \right)=
\frac{\pi}{3} 2.5 d \left( .5^2 d^2+.5 \cdot .35 d^2+.35^2 d^2 \right)=
1.43 d^3.
\ee \end{linenomath}
The volume of the cylindrical body of height $4.2\ d$ and radius $0.35\ d$ is
\begin{linenomath} \be
\label{4.02}
\pi H R^2=\frac{\pi}{4}(4.2d)(.7 d) ^2=1.62 d^3.
\ee \end{linenomath}
The volume for LV--Haack nose is derived from the cone equations \cite{cones} as $0.56 \pi L R^2$, where $L$ is the cone length and $R$ is the cone radius.  Thus, in our case the cone volume is $0.50 d^3$.  The total volume of the projectile is $3.55 d^3$.  If the average density of the projectile is $\rho$, then its total mass is
\begin{linenomath} \be
\label{4.03}
M_p=3.55 d^3 \rho.
\ee \end{linenomath}

The projectile of our choice has caliber $d=7.5\ cm$, density $\rho=11\ g/cm^3$, length 67.5 $cm$, and weight of 16.5 $kg$.  A 70 $kg$ warhead compartment can carry up to 3 such projectiles.
Using RASAero software \cite{rasaero} we have calculated the drag coefficient for the HPV at different Mach numbers and presented in Table \ref{T08} below.

\begin{center}
\begin{tabular}{|c|c|c|c|c|c|c|c|c|c|c|c|c|c|c|c|c|c|c|}
  \hline
  $M$   &.25   &.50   &.75   & 1.00  &1.25  & 1.50  & 1.75 & 2.00  & 2.25 & 2.50  & 2.75 \\
  \hline
  $C_d$ & .184 & .182 & .181 & .264 & .297 & .268  & .242 & .219  & .200 & .184 & .169  \\
  \hline
\end{tabular}\\
\begin{tabular}{|c|c|c|c|c|c|c|c|c|c|c|c|c|c|c|c|c|c|c|}
  \hline
  $M$   & 3.00 & 3.25 & 3.50 & 3.75 & 4.00 &4.25  &4.50  &4.75  & 5.00  &5.25  & 5.50\\
  \hline
  $C_d$ & .156  & .145 & .135 & .126 & .119 & .112 & .106 & .100  & .095  & .091 & .086 \\
  \hline
\end{tabular}\\
\begin{tabular}{|c|c|c|c|c|c|c|c|c|c|c|c|c|c|c|c|c|c|c|}
  \hline
  $M$     & 5.75 & 6.00  & 6.25 & 6.50  & 6.75 & 7.00 & 7.25 & 7.05 & 7.75 & 8.00 \\
  \hline
  $C_d$ & .081 & .076  & .072  & .069 & .066 & .062 & .060 & .058 & .056 & .055 \\
  \hline
\end{tabular}
\captionof{table}{Drag coefficient for Europrojectile \label{T08}}
\end{center}

We input the results of Rocket.m simulations and the drag coefficient tabulated in Table \ref{T06} above into Impact.m.  The program Impact.m calculates the projectiles' ranges, impact velocities, and descent drag loss.
The results of calculations are presented in Table \ref{T09} below.
The first row is the initial velocity.
The second row is the firing altitude.
The third row is the rocket firing time.
The fourth row is the apogee altitude.
The fifth row is the apogee velocity.
The sixth row is the warhead horizontal range from the firing point.
The seventh row is the impact velocity.
The eighth row is the descent drag loss.

\begin{center}
\begin{tabular}{|l|r|r|r||r|r|r||r|r|r|r|r|r|r|r|r|r|r|r|r|r|r|r|r|r|r|}
  \hline
  DLSRR $v_0$, $m/s$             &  500  &   600 &   700  &  500  &   600 &   700  &  500  &   600 &   700  \\
  DLSRR $h_0$, $km$              &   12  &    16 &    20  &   12  &    16 &    20  &   12  &    16 &    20  \\
  DLSRR $t_f$, $s$               &   30  &    30 &    30  &   50  &    50 &    50  &   80  &    80 &    80  \\
  \hline
  \hline
  Apogee $h$, $km$               &    93 &   116 &    133 &    92 &   110 &   122  &   77  &    95 &   111 \\
  Apogee $v$, $m/s$              & 1,239 & 1,310 &  1,399 & 1,170 & 1,265 & 1,398  & 1,151 & 1,241 & 1,349 \\
  \hline
  \hline
  Range, $km$                    &   328 &   386 &   442  &   305 &   361 &   418  &   264 &   319 &   374 \\
  Impact $v$, $m/s$              & 1,555 & 1,730 & 1,879  & 1,507 & 1,675 & 1,817  & 1,400 & 1,573 & 1,726 \\
  $\triangle v_{_{\text{loss}}}$, $m/s$  &   277 &   266 &   260  &   272 &   265 &   268  &   282 &   274 &   272 \\
  \hline
\end{tabular}
\captionof{table}{Europrojectile warhead performance \label{T09}}
\end{center}

From the above table we see that the Europrojectile has high impact velocity.  At such velocity, the projectile can inflict considerable damage due to kinetic energy.  One type of kinetic energy projectile which inflicts high area damage is a flechette projectile.
This projectile explodes at a distance of about a hundred meters from the target releasing thousands of tungsten flechettes weighing 1 $g$ to 2 $g$.  For a projectile with high impact velocity, the flechettes cause extensive damage \cite{Flechette1}.

\section{Aerodynamic Heating and Thermal Protection}
An object moving through atmosphere at supersonic velocity experiences two types of aerodynamic heating.  The first type is shock wave heating.  The supersonic object produces a shock wave in front of it.
Air temperature behind a shock wave can be as high as 20,000 $^o$C, which has been experienced by the Galileo probe when entering the Jupiter's atmosphere [60, p.9].  For the rockets we are considering the temperature of the shocked air will never exceed ``mere" 1,850 $^o$C, but this temperature is still high.

The second type is skin friction heating.  This is due to the high-speed friction of the rocket nose and sides against supersonic or hypersonic stream.  Generally, this friction contributes more to overall dynamic heating than the shock wave.

In this work, we use simplified expressions to find an upper bound of heat flux.  Exact calculation of heat flux requires extensive calculations using Aerodynamic Theory.  Designing an actual thermal protection requires not only advanced theoretical and computational work, but also many laboratory experiments.  Most rockets take thousands of expert-years to design \cite{SLVD}.

The expressions themselves do not differentiate between heat flux caused by the shock wave and heat flux caused by friction.  The heat transfer equations fall into two different classes.  These classes represent two types of flow -- laminar and turbulent. In this Section we present the expressions for heat flux generated by laminar and turbulent flows.

At the nose of the vehicle, the flow is always laminar.  At a short distance from the nose, it becomes turbulent and stays turbulent for the rest of the vehicle length.  Generally, turbulent flow transfers a higher heat flux to the wall then a laminar flow.

The expressions used to estimate laminar and turbulent heat fluxes on different types of surfaces have been obtained over decades as mathematical fits to experimental data.  These expressions may or may not have deeper physical meanings.  Some data patterns may have more than one expression providing a good fit \cite{nose,TMD,Aerodynamics,nose1,nose2,nose3}.

One of the main advantages of DLSRR over conventional long-range ballistic missiles is the relatively low heating of DLSRR during ascent.  The DLSRR reaches high velocities only in significantly rarified atmosphere.  Thus, the heat flux and heat load are considerably lower than those experienced by a conventional long-range rocket.

\subsection{General Expressions for Heat Flux}
As a rocket flies through the air at high speed, it experiences aerodynamic heating.
The rocket nose experiences laminar flow heating, while the nose cone and rocket cylinder experience turbulent flow heating.  Even though some parts of the nose cone may experience laminar flow at high altitudes, turbulent flow heating rate still provides a valid upper bound.
In this Subsection we present general expressions for the heat flux experienced by different points on a rocket surface during ascent.

Recovery temperature is the temperature the surface would have if it lost zero energy by thermal radiation or inward conduction.  Recovery temperature is such that air at that temperature has enthalpy $h_{_{aw}}$.  Adiabatic wall enthalpy, $h_{_{aw}}$ is given by
  \begin{linenomath} \be
  \label{5.01}
  h_{_{aw}}=h_{_a}+r_{_f}\ \frac{v^2}{2}.
  \ee \end{linenomath}
where $h_{_a}$ is the ambient air enthalpy, $v$ is the vehicle velocity, and $r_{_f}$ is the recovery factor \cite[p.9]{nose}.  The recovery factor is a dimensionless number, which is $1$ at a stagnation point, and $0.9$ for a turbulent boundary flow \cite[p.100]{TMD}.  A crude approximation for the recovery temperature is given by \cite[p.100]{TMD}:
  \begin{linenomath} \be
  \label{5.02}
  T_r=T_a \left( 1+.2 r_{_f} \mathcal{M}^2 \right),
  \ee \end{linenomath}
where $T_r$ is the recovery temperature, $T_a$ is the ambient air temperature.

\subsubsection{Stagnation Point Heat Flux}

\emph{The stagnation point} is a point in a flow field where the local velocity of the fluid is zero \cite[p.17]{Aerodynamics}.  The nose tip is the only stagnation point in the flow around a rocket.  The laminar heat flux to the stagnation point can be approximated by \cite[p.6,  Eq. (40)]{nose}:
  \begin{linenomath} \be
  \label{5.03}
  \dot{Q}_{_S}^{^{\text{laminar}}}=\dot{Q}_{_{K}}\ \frac{W}{m^2}\  \sqrt{\frac{\rho}{r}}\ \ v^3\ \ \
  \frac{h_{_{aw}}-h_{_w}}{h_{_{aw}}-h_{_a}},
  \ee \end{linenomath}
where $Q_{_S}$ denotes the total thermal energy absorbed by the stagnation point, and  $\dot{Q}_{_S}$ is its time derivative.
In Eq. (\ref{5.03}) above,
$\rho$ is the air density measured in $kg/m^3$,
$r$ is the nose radius in meters,
$h_{_w}$ is the specific enthalpy of the air at the wall,
$h_{_{aw}}$ is the specific enthalpy of the air at a hypothetical adiabatic wall,
$h_{_a}$ is the specific enthalpy of the ambient air,
$v$ is the rocket velocity in $m/s$ with respect to air and
  \begin{linenomath} \be
  \label{5.05}
  \dot{Q}_{_{K}}=1.83 \cdot 10^{-4}\ \frac{W}{m^2}
  \ee \end{linenomath}
is a constant.  Sutton and Graves give a similar expression \cite{nose1} with
  \begin{linenomath} \be
  \label{5.06}
  \dot{Q}_{_{K}}=1.74 \cdot 10^{-4}\ \frac{W}{m^2}
  \ee \end{linenomath}
According to Chapman \cite{nose2},
  \begin{linenomath} \be
  \label{5.07}
  \dot{Q}_{_{K}}=1.63 \cdot 10^{-4}\ \frac{W}{m^2}
  \ee \end{linenomath}
Detra and Hidalgo give a velocity dependent expression \cite{nose3},
  \begin{linenomath} \be
  \label{5.08}
  \dot{Q}_{_{K}}=1.45 \cdot 10^{-4}\ \frac{W}{m^2}\ V_{_{\text{kps}}}^{0.15},
  \ee \end{linenomath}
where $V_{_{\text{kps}}}$ is the rocket velocity in kilometers per second.  In this work, we use \\ $\dot{Q}_{_{K}}=1.83 \cdot 10^{-4}\ W/m^2$, which is the highest heating rate.
By including the blackbody radiation emanating from the rocket, we obtain the total heat flux passing through the stagnation point:
  \begin{linenomath} \be
  \label{5.09}
  \dot{Q}_{_S}^{^{\text{laminar}}}=\dot{Q}_{_{K}}\  \sqrt{\frac{\rho}{r}}\ \
  v^3
  \ \frac{h_{_{aw}}-h_{_w}}{h_{_{aw}}-h_{_a}}
  -e_{_w} \sigma_B T_w^4,
  \ee \end{linenomath}
where $\sigma_B=5.67 \cdot 10^{-8}\  W/m^2 K^4$ is the Stefan--Boltzmann constant, and $e_{_w}<1$ is the wall emissivity.

\subsubsection{General Turbulent Flow Heat Flux}
Turbulent flow heat flux is generally 3 to 6 times greater than laminar flow heat flux.  The hypersonic flow around any rocket almost always starts out laminar and then turns turbulent.  The transition from laminar to turbulent flow occurs at Reynolds number
$10^5 \le Re \le 10^6$ \cite[p.19]{LTT}.  The Reynolds number is defined in Equations (\ref{5.12}) and (\ref{5.14}) below.  The flow is turbulent over almost all surface area of the hypersonic vehicles we are considering in this paper.

Below we estimate the skin friction heat flux for general wall temperature and inclination.
According to \cite[p.iii]{nose}, the skin friction heat transfer is defined in terms of Stanton number,
  \begin{linenomath} \be
  \label{5.10}
  \dot{Q}^{^{\text{turbulent}}}=
  St \cdot \rho\ v \ \Big(h_{_{aw}}-h_{_w}\Big)=
  St \cdot \rho\ v \ \Big(h_{_{aw}}-h_{_a}\Big)
  \ \frac{h_{_{aw}}-h_{_w}}{h_{_{aw}}-h_{_a}}.
  \ee \end{linenomath}
The enthalpy supplied to the air by the shock wave travelling in front of the moving rocket is equal to $v^2/2$.  Thus,
  \begin{linenomath} \be
  \label{5.11}
  h_{_{aw}}-h_{_a}=\frac{v^2}{2}.
  \ee \end{linenomath}
Substituting (\ref{5.11}) into (\ref{5.10}), we obtain
  \begin{linenomath} \be
  \label{5.12}
  \dot{Q}^{^{\text{turbulent}}}
  =St \cdot \rho\ v \ \big(h_{_{aw}}-h_{_a}\big)
  \  \frac{h_{_{aw}}-h_{_w}}{h_{_{aw}}-h_{_a}}
  =St \ \frac{\rho\ v^3}{2}
  \ \frac{h_{_{aw}}-h_{_w}}{h_{_{aw}}-h_{_a}}
  .
  \ee \end{linenomath}

The Stanton number at distance $r y$ from the leading edge can be expressed in terms of skin friction coefficient \cite[p.305]{AHeat05}:
  \begin{linenomath} \be
  \label{5.13}
  St(r y)= \frac{C_f(r y)/2}{1+13 \big(Pr^{2/3}-1\big) \sqrt{C_f(r y)/2}},
  \ee \end{linenomath}
where $C_f(r y)$ is the skin friction coefficient at distance $r y$ from the leading edge and $Pr$ is the Prandtl number.  The Prandtl number for air is at least $0.71$.  The skin friction coefficient for DLSRR, under conditions we are considering, does not exceed $0.008$.  Substituting these values into (\ref{5.13}) we obtain
  \begin{linenomath} \be
  \label{5.14}
  St(r y) \le 0.60\ C_f(r y).
  \ee \end{linenomath}
Substituting (\ref{5.14}) into (\ref{5.12}), we obtain
  \begin{linenomath} \be
  \label{5.15}
  \dot{Q}^{^{\text{turbulent}}}
  =0.30\ C_f(r y) \ \rho\ v^3.
  \ee \end{linenomath}

In order to calculate the skin friction coefficient, we introduce the Reynolds number.  The Reynolds number for a point on a rocket body is given by
\cite[p.6]{AHeat01}
  \begin{linenomath} \be
  \label{5.16}
  Re=\frac{\rho v x}{\mu},
  \ee \end{linenomath}
where $x$ is the distance of a point from the stagnation point or the leading edge.The dynamic viscosity of the air is \cite[p.20-21]{AirVisc}:
  \begin{linenomath} \be
  \label{5.17}
  \mu \lessapprox 1.7 \cdot 10^{-5}\ \frac{kg}{m \ s}\ \left( \frac{T}{240\ ^oK} \right)^{0.7}.
  \ee \end{linenomath}
The temperature of 240 $^o$K is close to the ambient air temperature in the region of DLSRR paths.  Combining (\ref{5.16}) and (\ref{5.17}), we obtain the Reynolds number for any point on the rocket surface:
   \begin{linenomath} \be
  \label{5.18}
  Re \approx 70 \cdot 10^6
             \left(\frac{\rho}{1\ kg/m^3}\right)
             \left( \frac{v}{1\ km/s}    \right)
             \left( \frac{x}{1\ m}       \right).
  \ee \end{linenomath}
The above equation is valid for ambient air temperature for most maneuvers.  The effect of heating of air by friction and shock wave is described below.

The skin friction coefficient is very well approximated by \cite[p.4]{SkinFr02}:
   \begin{linenomath} \be
   \label{5.19}
   C_f=0.295\ \frac{T_a}{T^*}\ \left[ \log \left(Re\ \frac{T_a}{T^*}\ \frac{\mu_a}{\mu^*} \right) \right]^{-2.45},
   \ee \end{linenomath}
where
$T_a$ is the ambient air temperature,
$T^*$ is the air temperature at the boundary,
$\mu_a$ is the ambient air viscosity,
and $\mu^*$ is the viscosity at the boundary.  Expression (\ref{5.19}) shows excellent agreement with experiment \cite[p.8]{SkinFr02}.
The temperature at the boundary is the temperature corresponding to the enthalpy $h^*$, which is \cite[p.6]{AHeat01}:
   \begin{linenomath} \be
   \label{5.20}
   h^*=0.22\ h_{_{aw}}+0.50\ h_{_w}+ 0.28\ h_{_a},
   \ee \end{linenomath}
where
$h_{_{aw}}$ is the adiabatic enthalpy,
$h_{_w}$ is the enthalpy of air at wall temperature, and
$h_{_a}$ is the enthalpy of ambient air.
The term $T_a/T^*$ in Eq. (\ref{5.19}) is due to the fact that the air density is inversely proportional to the air temperature, and the friction force is directly proportional to the air density.

From (\ref{5.17}), air viscosity grows approximately as temperature to the power 0.7.    Thus,
   \begin{linenomath} \be
   \label{5.21}
   \begin{split}
   C_f&=0.295\ \frac{T_a}{T^*}\ \left[ \log \left(Re\ \frac{T_a}{T^*}\ \frac{\mu_a}{\mu^*} \right) \right]^{-2.45}
   \le 0.295\ \frac{T_a}{T^*}\ \left[ \log \left(Re\ \left(\frac{T_a}{T^*}\right)^{1.7} \right) \right]^{-2.45}\\
   &=0.295\ \frac{T_a}{T^*}\ \left[\log Re+1.7 \log \left(\frac{T_a}{T^*}\right)  \right]^{-2.45}
   =\frac{0.295\ \left[\log Re-1.7 \log \left(T^*/T_a\right)  \right]^{-2.45} }{T^*/T_a}.
   \end{split}
   \ee \end{linenomath}
The range of validity for Eq. (\ref{5.21}) above is wide enough to describe the conditions for all ascending rockets.
In Equation (\ref{5.21}) above, $\log$ represents log base 10.  We express (\ref{5.21}) in terms of natural logarithm:
   \begin{linenomath} \be
   \label{5.22}
   C_f
   =\frac{2.28\ \left[\ln Re-1.7 \ln \left(T^*/T_a\right)  \right]^{-2.45} }{T^*/T_a}.
   \ee \end{linenomath}
Combining (\ref{5.22}) and (\ref{5.15}), we obtain
  \begin{linenomath} \be
  \label{5.23}
  \dot{Q}^{^{\text{turbulent}}}
  =\frac{0.68\ \left[\ln Re-1.7 \ln \left(T^*/T_a\right)  \right]^{-2.45} }{T^*/T_a} \ \rho\ v^3
  \ \frac{h_{_{aw}}-h_{_w}}{h_{_{aw}}-h_{_a}}.
  \ee \end{linenomath}

For the rockets and velocities we are considering, the Reynolds number varies between $1 \cdot 10^6$ and $1 \cdot 10^8$.  The quotient $T^*/T_a$ varies between 1 and 8.  For all values of $Re$ and $T^*/T_a$ considered, $C_f$ is strongly and strictly decreasing function of $T^*/T_a$.  Hence, skin friction coefficient and turbulent heating are strongly decreasing with rising wall temperature.

\subsubsection{Rocket Cylinder}
For the rocket cylinder of DLSRR, the turbulent flow heat flux is calculated by (\ref{5.23}) with the Reynolds number $Re$ given by (\ref{5.18}).  At the base of the rocket cylinder, the distance from the stagnation point is $x=1\ m$.  Substituting this distance into (\ref{5.18}), we obtain the Reynolds number for the leading edge:
  \begin{linenomath} \be
  \label{5.24}
  Re \approx 70 \cdot 10^6
             \left(\frac{\rho}{1\ kg/m^3}\right)
             \left( \frac{v}{1\ km/s}    \right).
  \ee \end{linenomath}
For further parts of the rocket cylinder, the Reynolds number is higher and the heat flux is a little lower.

\subsubsection{Nose Cone}
Conical shape surface experiences 1.3 times the heat flux experienced by a flat plate \cite[p. 11]{AHeat03}.  Another factor leading to an increase in turbulent flow heating rate on a cone relative to cylinder wall is the fact that the air pressure on the cone will exceed ambient atmospheric pressure.  Any shape moving through the air stream with Mach number $\mathcal{M}$ having tangential angle $\theta$ to the air stream will incur a dynamic pressure of
  \begin{linenomath} \be
  \label{5.25a}
  P_{_d}
  =P\ \frac{\gamma}{2}\ \mathcal{M}^2\ \big(\sin \theta \big)^2,
  \ee \end{linenomath}
where $P$ is the ambient atmospheric pressure.
We calculate the total pressure exerted on an inclined surface in an air stream by adding ambient pressure and dynamic pressure.
The total pressure will exceed the ambient pressure by a factor of
  \begin{linenomath} \be
  \label{5.25}
  \frac{P_{_w}}{P}
  =1+\frac{\gamma}{2}\ \mathcal{M}^2\ \big(\sin \theta \big)^2
  \le 1+6\cdot 10^{-6}\ \big(\sin \theta \big)^2 v^2,
  \ee \end{linenomath}
where $v$ is the rocket velocity in $m/s$, and $\theta$ is the angle between the cone surface and the oncoming air stream .  The heat transfer rate given in (\ref{5.23}) will be multiplied by the same amount.  Thus,
  \begin{linenomath} \be
  \label{5.26}
  \begin{split}
  \dot{Q}^{^{\text{turbulent}}}_{_{\text{cone}}}
  &\le \Big( 1+6\cdot 10^{-6}\ \big(\sin \theta \big)^2 v^2\Big)\\
  &\times \frac{0.88\ \left[\ln Re-1.7 \ln \left(T^*/T_a\right)  \right]^{-2.45} }{T^*/T_a} \ \rho\ v^3
  \ \frac{h_{_{aw}}-h_{_w}}{h_{_{aw}}-h_{_a}}
  ,
  \end{split}
  \ee \end{linenomath}
where
$T^*$ is the temperature for which air enthalpy is given by (\ref{5.20}),
$T_a$ is the air temperature,
$T_w$ is the rocket wall temperature,
and $Re$ is the Reynolds number given by (\ref{5.18}).


Using the above information about the aerodynamic heating we have written a MatLab program \textbf{ThermalAnalysis.m} described in Appendix.  This program calculates the temperatures of DLSRR stagnation point, several points on DLSRR nose cone, and rocket cylinder base during the powered portion of DLSRR ascent.  The program takes in the time series data containing ambient air temperature, ambient air density, and rocket velocity.  These time series are produced by the program Rocket.m, which is also described in Appendix.

Equation (\ref{5.09}) is used to calculate stagnation point heat flux.
Equation (\ref{5.26}) is used to calculate the heat flux on the rocket nose cone caused by turbulent flow.
All of the aforementioned fluxes are functions of the local rocket wall temperature.
The program finds wall temperatures at the stagnation point and several points on rocket nose such that the absorbed heat flux is equal to the heat flux radiated away.
These temperatures are determined for all time intervals during the rocket's powered ascent.

Equation (\ref{5.23}) is used to calculate the heat flux on the rocket cylinder base caused by the turbulent flow.
The cylinder does not radiate away any of absorbed heat.  Being a heat sink shield, it absorbs the heat and gets hotter.  The program calculates the cylinder base temperature at the end of the powered flight.

The program ThermalAnalysis.m performs calculations for rocket heating.  The results of these calculations are presented in Subsection 5.3.

\subsection{Thermal Protection Systems}

There are three categories of heat shields.  These are radiative shields, ablative shields, and heat sink shields.  Radiative shields, which radiate away almost all of the heat flux they receive. They conduct almost no heat.  DLSRR nose cone uses a radiative heat shield.  Ablative shields absorb the heat flux.  The heat is dissipated by pyrolysis and sublimation of the shield material \cite[p.80]{Shields}.  DLSRR does not use ablative heat shields anywhere.  Heat sink shields absorb heat flux.  These shields generally consist of light metals which can absorb considerable thermal energy without becoming extremely hot.  DLSRR cylinder rocket uses Aluminum 6061 T6 cylinder wall as a "inherent" heat shield.  During the rocket flight the wall experience heating, but not enough to become dysfunctional.

\subsubsection{Radiative Heat Shields}
A typical radiative heat shield consists of three layers.  The outermost layer is composed of a thin metal sheet.  The metal should be resistant to oxidation and have high emissivity.  Emissivity is the ``ratio of the radiant flux emitted per unit area to that of an ideal black body at the same temperature" \cite[p.2-45]{crc}.  The next layer is flexible thermal insulation composed of ceramic fiber.  The innermost layer is the underlying rigid structure.  This is the shield used on DLSRR nose cone.  In some works, the whole three layer shield is called a metal heat shield \cite{NiChr}.\\ \\
\textbf{Outermost Layer -- Oxidation Resistant Metals}\\
The outermost layer of the metal heat shield is subject to both high heat flux and hypersonic oncoming wind.  This layer must have high melting point and high oxidation resistance.  Oxidation resistance of material is defined in terms of its recession rate under the influence of high temperature oxidizing environment.  The recession rate depends on material, temperature, air density, and oncoming wind velocity.  In some cases, recession follows a linear law:
 \begin{linenomath} \be
 \label{5.27}
 d_{_r}(t)=K_{_{\text{linear}}}t,
 \ee \end{linenomath}
where $d_{_r}(t)$ is the recession distance, $t$ is the exposure time, and $K_{_{\text{linear}}}$ is a linear recession constant.  The constant $K_{_{\text{linear}}}$ has units of $m/s$.    Some oxidation-resistant metals are protected from oxidation by an outer oxide layer \cite{NiOx}.  This layer is protective for temperatures beyond the oxide melting point.  For instance, refractory metals such as tungsten and molybdenum have poor oxidation resistance due to the low melting points of their corresponding oxides.  In cases where metal oxide does provide protection, metal recession follows a parabolic law:
 \begin{linenomath} \be
 \label{5.28}
 d_{_r}(t)=\left( K_{_{\text{parabolic}}}t \right)^{1/2},
 \ee \end{linenomath}
where $d_{_r}(t)$ is the recession distance, $t$ is the exposure time, and $K_{_{\text{parabolic}}}$ is a parabolic recession constant.  The constant $K_{_{\text{parabolic}}}$ has units of $m^2/s$.

For temperatures up to 1,300 $^o$C, Nickel has good oxidation resistance.  In static air at atmospheric pressure and 1,300 $^o$C temperature, it has parabolic oxidation rate with constant\\ $K_{_{\text{parabolic}}}=2.5 \cdot 10^{-15}\ m^2/s$ \cite{NiOx}.
Under these conditions, it takes 10,000 seconds to oxidize a layer of nickel 0.005 $mm$ thick.
Inconel's oxidation resistance is slightly lower.
In static air at atmospheric pressure and 1,300 $^o$C temperature, it has parabolic oxidation rate with constant $K_{_{\text{parabolic}}}=7 \cdot 10^{-14}\ m^2/s$ \cite[p.19]{IncOx}.
Nickel and Inconel with oxidised surface have emissivity of at least 0.75 \cite[p.E-386]{crc82}. 
Inconel 625 sheets 0.5 $mm$ thick are available at \$380 per $m^2$ \cite{Inconel1}.  These sheets have areal density of 4.2 $kg/m^2$.
An alloy containing 80\% Nickel and 20\% chromium has excellent properties and can be multiply reused as heat shield up to 1,200 $^o$C \cite[p.1]{NiChr}.  The oxidized surface of the alloy has emissivity of 0.7 to 0.9 \cite[p.184]{NiChr01}.  In static air at atmospheric pressure and 1,300 $^o$C temperature, the alloy has parabolic oxidation rate with constant $K_{_{\text{parabolic}}}<2 \cdot 10^{-15}\ m^2/s$ \cite[p.207]{NiChr01}.

Chromium is oxidation-resistant up to 1,550 $^o$C \cite{ChrOx00}.  A layer of chromium oxide protects chromium from rapid oxidation at high temperatures.  
Pure chromium should have been protected by its oxide up to its melting point of about 1,900 $^o$C.  Unfortunately, the Cr$_2$O$_3$ forms eutectic with lower oxides and melts away at 1,650 $^o$C \cite{ChrOx00}.
In air at atmospheric pressure and 1,550 $^o$C temperature, chromium has parabolic oxidation rate with constant $K_{_{\text{parabolic}}}=1 \cdot 10^{-12}\ m^2/s$ \cite{ChrOx02}. 

More expensive metals and alloys are needed to provide oxidation protection at higher temperatures \cite{RefractOx02}.   
For DLSRR nose cone, however, Nickel, Inconel, and Chromium are sufficient.\\ \\
\textbf{Second Layer -- High Temperature Thermal Insulation}\\
Relatively inexpensive ceramic felt and cloth thermal isolation is available.
CeramFiber thermal insulation works up to 1,260 $^o$C.
A sheet weighing 3 $kg/m^2$ costs \$58 per $m^2$ \cite{Fiber1}.
Superwool HT Paper 2 $mm$ thick has areal density of 0.45 $kg/m^2$, thermal conductivity below 0.25 $W/(m\ ^oK)$, and good properties up to 1,300 $^o$C \cite[p.46]{CeramicF1}.
Zirconia Grade ceramic fiber paper works up to 1,420 $^o$C.
A sheet weighing 3 $kg/m^2$ costs \$147 per $m^2$ \cite{Fiber2}.
All the aforementioned fibers can resist temperatures 100 $^o$C higher than their classification for about a minute.\\ \\
\textbf{DLSRR Thermal Insulation}\\
DLSRR nose cone has a simple radiative heat shield.  The stagnation point and the first 10 $cm$ of the cone are covered by a 1 $mm$ thick chromium sheet.  The second layer at that area consists of a 1 $cm$ thick Zirconia Grade ceramic fiber paper.  This combination can withstand temperatures of up to 1,520 $^o$C which is 1,793 $^o$K for duration of SRS ascent.  The rest of the nose cone is covered by a 0.5 $mm$ thick nickel or Inconel sheet.  The second layer consists of CeramFiber blanket 1 $cm$ thick.  This shield section can withstand temperatures of up to 1,260 $^o$C which is 1,533 $^o$K.

\subsubsection{Heat Sink Shields}
The heat sink shield consists of three layers similar to metal radiative heat shield.  The outermost layer is composed of metal.  The next layer is flexible thermal insulation composed of ceramic fiber.  The innermost layer is the underlying rigid structure.  The difference between heat sink shield and radiative heat shield is the function of the outer layer.  Whereas the outer layer of the radiative heat shield radiates away the heat flux, the outer layer of heat sink shield absorbs the thermal energy.  Heat sink shields are most useful for absorbing short and powerful heat fluxes.  These shields have been used for suborbital vehicles since 1950s \cite[p.23]{Home}.

Inherent heat sink shield is a part of a rocket or aircraft which has another primary function and also fulfils an extra function as a heat shield.  For instance, a military aircraft may have armor which can act as a heat sink shield during short-time supersonic maneuvers.  A solid rocket's metal shell mainly serves to contain the pressure of fuel grain combustion product.  This shell may serve as a inherent heat sink shield as long as it does not overheat during hypersonic flight.

DLSRR has such inherent heat sink shield.  As we mentioned in Section 4.1, the rocket cylinder consists of a 0.5 $cm$ thick aluminum wall.  Below we calculate the heat load such wall can withstand.

Rocket cylinder wall can be heated only so far as it retains tensile strength to hold the solid rocket motor.  Aluminum 6061 T6 is the material composing the artillery rocket cylinder wall.  Its yield strength is 283 $MPa$ at -28 $^o$C, 276 $MPa$ at 24 $^o$C, 262 $MPa$ at 100 $^oC$, 250 $MPa$ at 118 $^oC$, and 214 $MPa$ at 149 $^oC$.  On further heating, it rapidly loses strength \cite{Al6061T6v2}.

It takes 73 $J/g$ to heat aluminum from 20 $^o$C to 100 $^o$C \cite[p. 12-190]{crc}.  The 0.5 cm thick aluminum wall has an areal density of 13.5 $kg/m^2$.  During the rocket motor burning, it can absorb a heat flux of
  \begin{linenomath} \be
  \label{5.29}
  73\ \frac{J}{g} \times 1.35 \cdot 10^4\ \frac{g}{m^2}=
  9.9 \cdot 10^5\ \frac{J}{m^2}.
  \ee \end{linenomath}
As we show in Subsubsection 5.3.3, the heat flux at cylinder base is under 1 $MJ/m^2$.

Long-range rockets fired from the ground sustain an order of magnitude higher heat load.  These rockets must have extensive outer heat shields.  This is one of the reasons such rockets are expensive
\cite[p. 103-106]{TMD}.

\subsection{DLSRR Heating}
In this subsection we estimate the maximal heat load on various parts of the artillery rocket (DLSRR) during its' ascent.  The 30 $cm$ DLSRR is described in great detail in Section 3.4.  In this Subsection we describe the heating of three parts of DLSRR: the stagnation point, the nose cone and the rocket cylinder.  The nose cone experiences greater heat flux than the rocket cylinder.
The steps in heat flux calculations for the three parts are somewhat different.

The three parts of DLSRR have different heat shields.  Both DLSRR stagnation point and DLSRR nose cone use a radiative heat shield described in Subsection 6.3.  The temperature of this shield at any moment depends on the local heat flux.  DLSRR cylinder uses an inherent heat sink shield described in Subsection 6.3.  The temperature of this shield grows slowly almost in proportion to the absorbed heat load.  DLSRR outer shell containing the three aforementioned parts is presented below:
\begin{center}
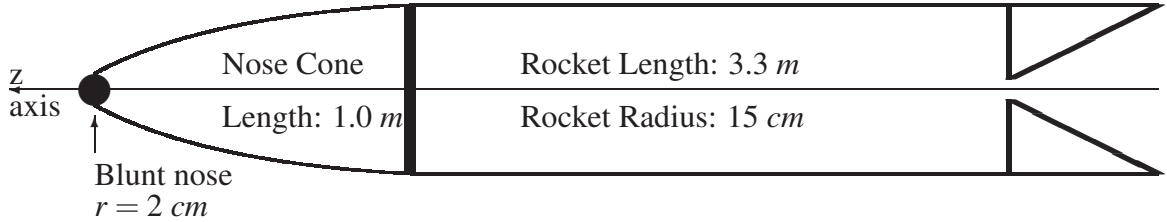

\setlength{\unitlength}{1.4mm}
\begin{picture}(100, 20)
\linethickness{0.3mm}

\put(30,2){\line(1,0){70}}
\put(30,18){\line(1,0){70}}
\multiput(30,2)(-.2,0){5}{\line(0,1){16}}
\put(30,2){\line(0,1){16}}
\qbezier(0,8.5)(10,3)(30,2)
\qbezier(0,11.5)(10,17)(30,18)

\put(0,10){\circle*{3}}
\linethickness{0.1mm}
\put(0,4){\vector(0,1){3.5}}
\linethickness{0.3mm}
\put(0,1){Blunt nose}
\put(0,-2){$r=2\ cm$}

\put(12,11.5){Nose Cone}
\put(12,6.5){Length: 1.0 $m$}

\put(40,11.5){Rocket Length: 3.3 $m$}
\put(40,6.5){Rocket Radius: 15 $cm$}

\multiput(86.5,11)(-.1,0){10}{\line(2,1){14}}
\multiput(86.5,9)(-.1,0){10}{\line(2,-1){14}}

\multiput(86,11)(-.1,0){3}{\line(0,1){7}}
\multiput(86,9)(-.1,0){3}{\line(0,-1){7}}

\linethickness{0.1mm}
\put(100,10){\vector(-1,0){108}}
\put(-8,10.5){z}
\put(-8,7.5){axis}

\end{picture}
\captionof{figure}{DLSRRouter shell \label{F09}}
\end{center}
\subsubsection{DLSRR Stagnation Point}
During the rocket ascent, DLSRR stagnation point experiences laminar flow heating.  This heat is radiated away by oxidized chromium covering the stagnation point.  The emissivity of oxidized chromium surface is at least 0.6 \cite[p.E-387]{crc82}.  The total heat flux to the stagnation point given in (\ref{5.09}) is
  \begin{linenomath} \be
  \label{5.30}
  \dot{Q}_{_S}^{^{\text{laminar}}}=\dot{Q}_{_{K}}\  \sqrt{\frac{\rho}{r}}\ \
  v^3
  \ \frac{h_{_{aw}}-h_{_w}}{h_{_{aw}}-h_{_a}}
  -e_{_w} \sigma_B T_w^4 \le
  \dot{Q}_{_{K}}\  \sqrt{\frac{\rho}{r}}\ \
  v^3
  \ \frac{h_{_{aw}}-h_{_w}}{h_{_{aw}}-h_{_a}}
  -0.6\ \sigma_B T_w^4
  ,
  \ee \end{linenomath}
where $\sigma_B=5.67 \cdot 10^{-8}\  W/m^2 K^4$ is the Stefan--Boltzmann constant.
The upper bound for stagnation point temperature is calculated by setting the maximum heat flux to zero:
\begin{linenomath} \be
  \label{5.31}
  \dot{Q}_{_{K}}\  \sqrt{\frac{\rho}{r}}\ \
  v^3
  \ \frac{h_{_{aw}}-h_{_w}}{h_{_{aw}}-h_{_a}}
  -0.6\ \sigma_B T_w^4=0.
  \ee \end{linenomath}
Notice, that in the above Eq. (\ref{5.31}), the air enthalpy at wall temperature $h_{_w}$ is a function of the wall temperature $T_w$.

The program ThermalAnalysis.m uses Eq. (\ref{5.31}) to calculate stagnation point temperature. This calculation is done for every point on DLSRR trajectory.

\subsubsection{DLSRR Nose Cone}
During the rocket ascent, DLSRR nose cone experiences turbulent flow heating.
This heat is radiated away by oxidized Inconel covering the stagnation point.
The emissivity of oxidized Inconel surface is at least 0.75 \cite[p.E-387]{crc82}.   The total heat flux to the nose cone is obtained by subtracting the radiative heat flux from the convective heat flux given in (\ref{5.26}).   For reader's convenience, we provide the formula for total heat flux with the notations described below formula (\ref{5.26}).
  \begin{linenomath} \be
  \label{5.32}
  \begin{split}
  \dot{Q}^{^{\text{turbulent}}}_{_{\text{cone}}}
  &\le \Big( 1+6\cdot 10^{-6}\ \big(\sin \theta \big)^2 v^2\Big)\\
  &\times \frac{0.88\ \left[\ln Re-1.7 \ln \left(T^*/T_a\right)  \right]^{-2.45} }{T^*/T_a} \ \rho\ v^3
  \ \frac{h_{_{aw}}-h_{_w}}{h_{_{aw}}-h_{_a}}-0.75\ \sigma_B T_w^4.
  \end{split}
  \ee \end{linenomath}
The angle of inclination, $\theta$, towards the air stream is different for different parts of nose cone.
The upper bound for nose cone temperature is calculated by setting the maximum heat flux to zero:
  \begin{linenomath} \be
  \label{5.33}
  \begin{split}
  \Big( 1+6\cdot 10^{-6}\ \big(\sin \theta \big)^2 v^2\Big)
   \frac{0.88\  \rho\ v^3}{
   \frac{T^*}{T_a}\ \left[\ln Re-1.7 \ln \left(\frac{T^*}{T_a}\right)  \right]^{2.45}}
  \ \frac{h_{_{aw}}-h_{_w}}{h_{_{aw}}-h_{_a}}-0.75\ \sigma_B T_w^4=0.
  \end{split}
  \ee \end{linenomath}
Notice, that in the above Eq. (\ref{5.33}), both the air enthalpy at wall temperature $h_{_w}$ and boundary air temperature $T^*$ are functions of the wall temperature $T_w$.

In order to solve Eq. (\ref{5.33}), we must find the angle of inclination $\theta$ for different points on the nose cone.  We visualize the nose cone below:
\begin{center}
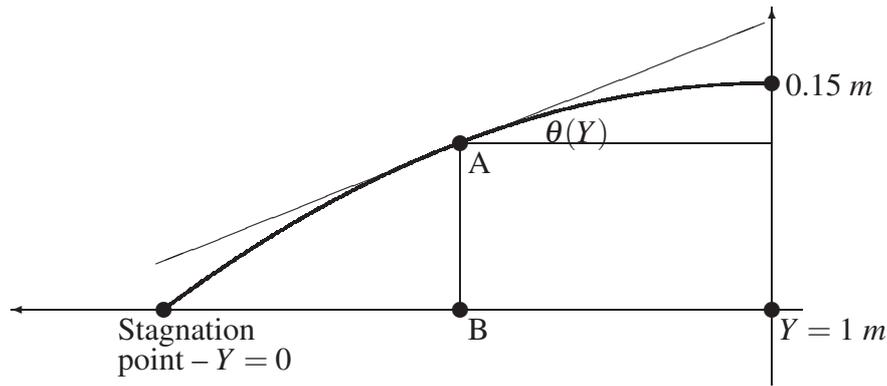

\setlength{\unitlength}{2mm}
\begin{picture}(50, 25)
\linethickness{0.3mm}
\qbezier(10,5)(30,20)(50,20)
\put(10,5){\circle*{1}}
\put(50,20){\circle*{1}}
\put(50.9,19.2){0.15 $m$}
\put(7,3){Stagnation}
\put(7,1){point -- $Y=0$}
\linethickness{0.1mm}
\put(52,5){\vector(-1,0){52}}
\put(50,0){\vector(0,1){25}}

\put(29.5,16){\line(1,0){20.5}}
\put(29.5,16){\line(0,-1){11}}
\put(29.5,16){\line(5,2){20}}
\put(29.5,16){\line(-5,-2){20}}
\put(35,16.1){$\theta(Y)$}

\put(50.5,3){$Y=1\ m$}
\put(50,5){\circle*{1}}
\put(29.5,5){\circle*{1}}
\put(29.5,16){\circle*{1}}
\put(30,3){B}
\put(30,14){A}
\end{picture}
\captionof{figure}{Nose cone \label{F10}}
\end{center}
Any point A on the cone can be located in terms of an angular coordinate $\phi \in [0, 2\ \pi )$ and a non-angular coordinate $Y$.  If B is the projection of A onto Y axis then Y is the distance between A and the stagnation point.  Heat flux and heat shield temperature are independent of $\phi$ due to the angular symmetry.  Below, we calculate the heat flux and the heat shield temperature for several values of $Y$.  For an ogive cone with radius 0.15 $m$ and length 1.0 $m$, the inclination of any plane segment at coordinate $Y$ is
  \begin{linenomath} \be
  \label{5.34}
  \theta(Y)\ \approx\ 0.3 \cdot \frac{1\ m-Y}{1\ m}.
  \ee \end{linenomath}

The program ThermalAnalysis.m uses Eq. (\ref{5.33}) to calculate nose cone temperatures at different points on the nose cone as functions of time.  This calculation is done for every point on DLSRR trajectory.  The heat shield temperatures for DLSRR rocket for different firing altitudes, initial velocities, and rocket firing times are shown in Figure \ref{F11} below.

\begin{flushleft}
\includegraphics[width=5cm]{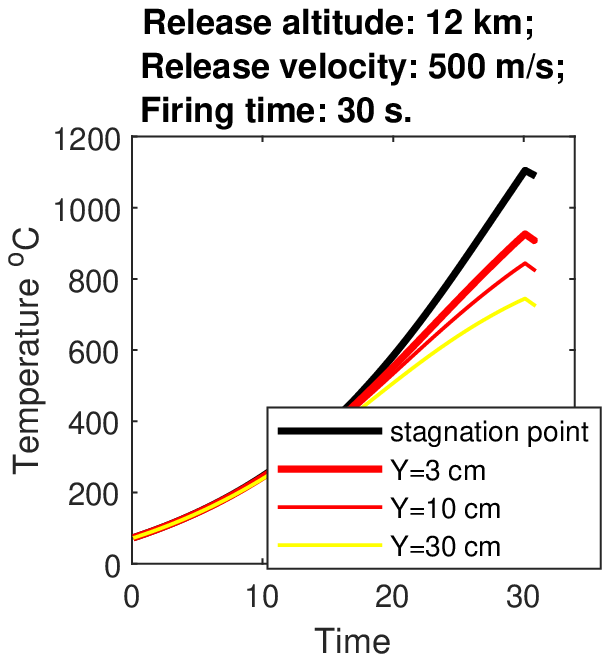}
\includegraphics[width=5cm]{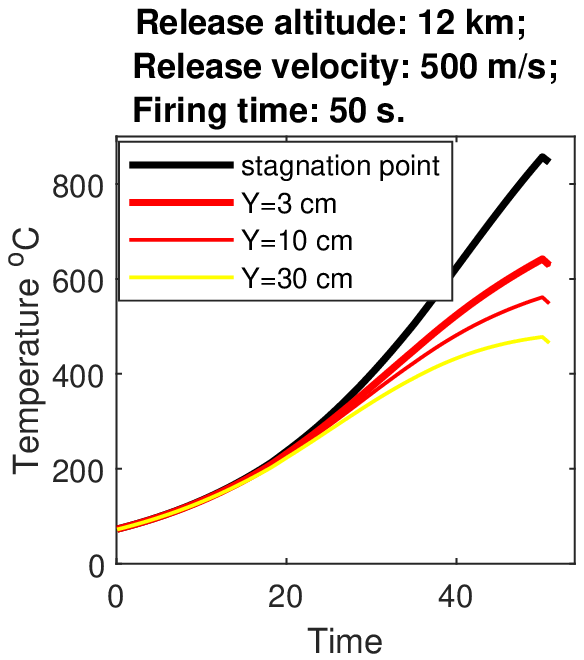}
\includegraphics[width=5cm]{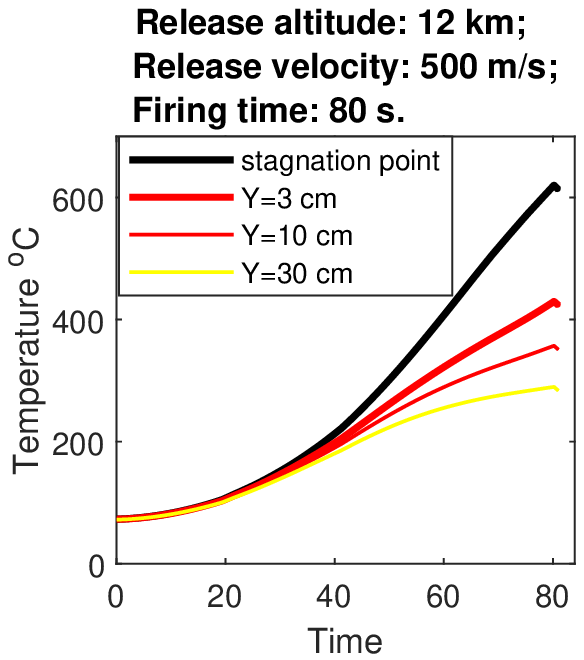}
\includegraphics[width=5cm]{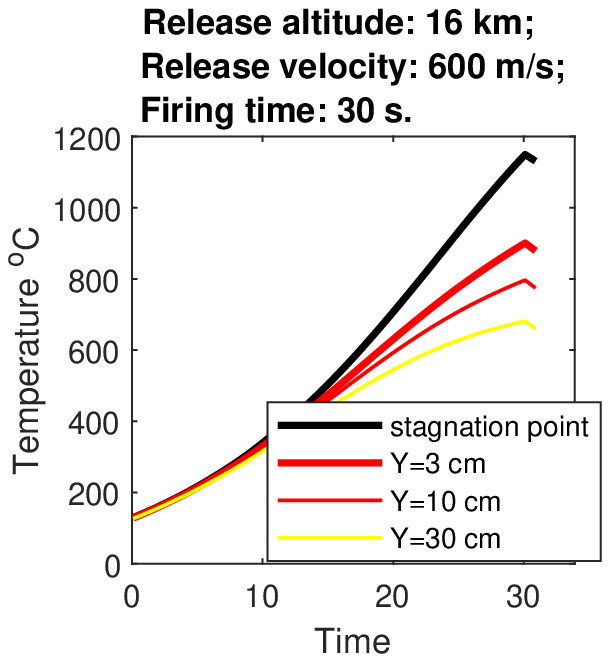}
\includegraphics[width=5cm]{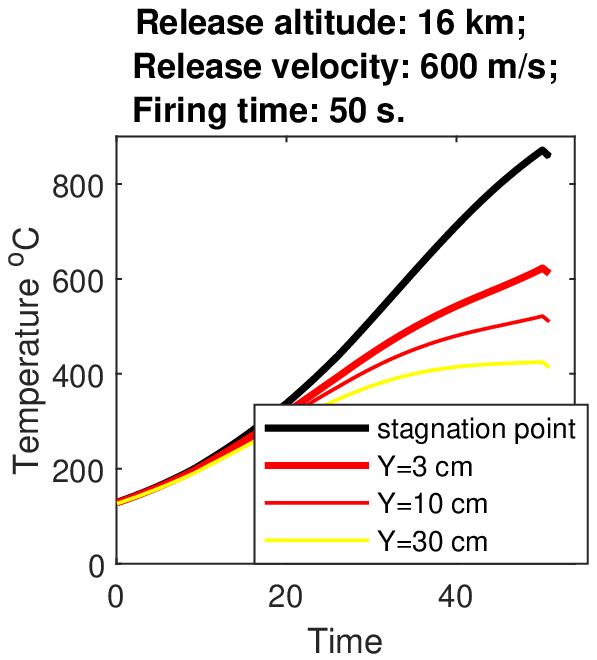}
\includegraphics[width=5cm]{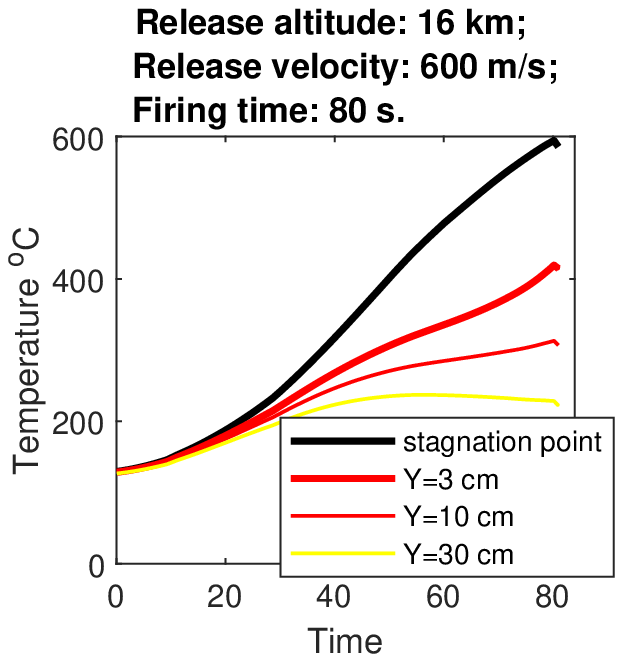}
\includegraphics[width=5cm]{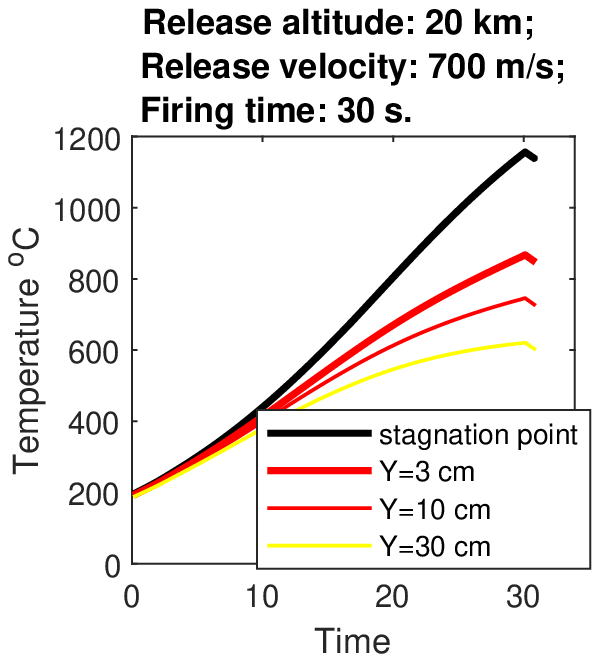}
\includegraphics[width=5cm]{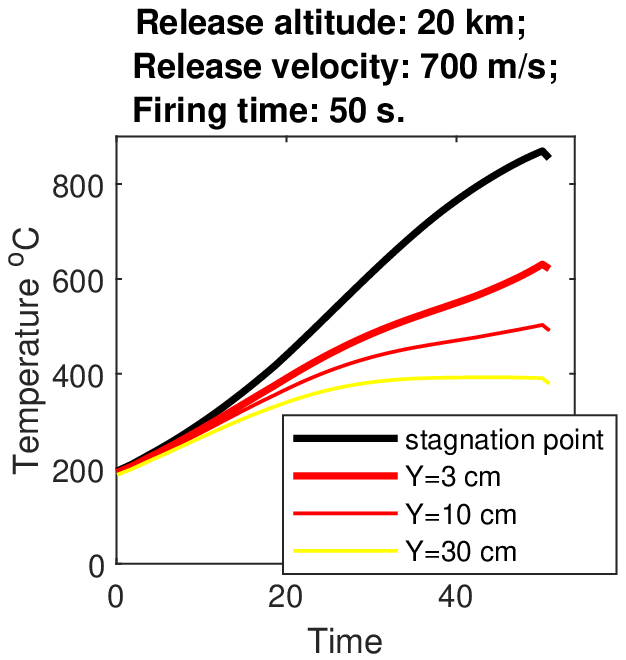}
\includegraphics[width=5cm]{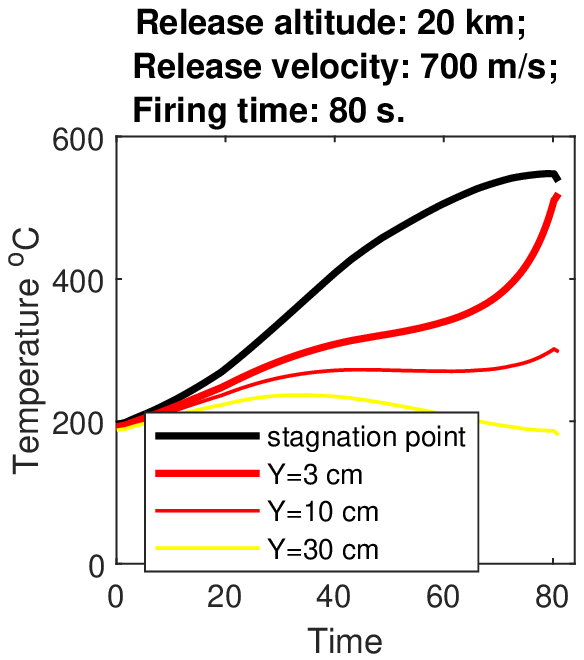}
\captionof{figure}{Aerodynamic heating of DLSRR nose cone \label{F11}}
\end{flushleft}

All of the maximum temperatures are within the capabilities of three-layered radiative heat shields described in Subsection 5.2.  The severest conditions arise when the rocket firing time is the shortest -- 30 $s$.  The maximum temperature of under 1,200 $^o$C is reached for a few seconds.  Not only chromium, but also nickel and Inconel can easily withstand such temperature.

The heat flux at the stagnation point is proportional to $\sqrt{\rho}$.
        The heat flux on the nose cone is proportional to $\rho$,
        where $\rho$ is the ambient air density.
        As we see on the last plot of the last row, the heat flux 3 $cm$ from stagnation point approaches the heat flux at the stagnation point as the rocket ascends.

\subsubsection{DLSRR Cylinder}

The turbulent flow heating rate is given by (\ref{5.26}).  All the heat is absorbed by the 0.5 $cm$ thick aluminum rocket body.  Almost no heat is radiated away.  Assuming DLSRR body starts out at 20 $^o$C, the program ThermalAnalysis.m calculates how much the rocket body heats up during the motor firing.  After firing, the rocket body may absorb more heat, but it will have no effect on rocket performance.  The temperature of the rocket cylinder base after firing is tabulated in Table \ref{T10}.
\begin{center}
  \begin{tabular}{|l|r|r|r|r|r|r|r|r|r|r|r|r|r|r|r|r|r|r|}
    \hline
  Firing time              &30 $s$   &40 $s$   &50 $s$  &60 $s$  &80 $s$  \\
    \hline
  $H=12\ km$, $v=500\ m/s$ & 94 $^o$C& 81 $^o$C&68 $^o$C&58 $^o$C&47 $^o$C\\
  $H=16\ km$, $v=600\ m/s$ & 75 $^o$C& 66 $^o$C&59 $^o$C&55 $^o$C&46 $^o$C\\
  $H=20\ km$, $v=700\ m/s$ & 61 $^o$C& 56 $^o$C&52 $^o$C&48 $^o$C&43 $^o$C\\
    \hline
  \end{tabular}
  \captionof{table}{Rocket cylinder base heating \label{T10}}
\end{center}
For combinations of rocket altitude, velocity, and firing time, rocket cylinder heating presents no problem.

\section{Conclusion}

In this work we have demonstrated the feasibility of drone launched short range rockets launched from a rocket launcher aircraft.  DLSRR deliver warheads to targets at a distance of 263 $km$ to 442 $km$ with impact velocity of 1,400 $m/s$ to 1,880 $m/s$.  RLA never approaches the target closer than 220 $km$ making it much less vulnerable to anti-aircraft weapons than modern fighters and bombers.

The cost of delivery of munitions by the DLSRR should be much lower than the cost of delivery of munitions by modern short range ballistic missiles (SRBM).  This is the case due to the greatly reduced requirements placed on the DLSRR compared to TBM.  First, DLSRR require much lower $\triangle v$ than SRBM.  Second, DLSRR can burn fuel over 30 $s$ to 80 $s$ which is much slower than most modern TBMs.  Third, as we have shown in Subsection 5.3, the DLSRR reach high velocity in high stratosphere and experience only moderate heating.  Stagnation point temperature never exceeds 1,200 $^o$C, and nose cone temperature never exceeds 960 $^o$C.  Such temperatures require relatively inexpensive heat protection.

A DLSRR having an initial mass of 300 $kg$ can deliver a 60 $kg$ charge to the target.  The charge can consist of one or more projectiles.  The ranges and impact velocities for a warhead containing four Europrojectiles are listed in Table \ref{T09}.  High release altitude and velocity combined with shorter burning times increase the warhead's range and impact velocity.  The same factors increase launch cost.

\appendix
\begin{center}
  {\Huge \textbf{Appendix}}
\end{center}

\section{Programs Used}
\begin{center}
\end{center}
\begin{description}
  \item[Rocket.m] performs calculations for the DLSRR.  The user inputs DLSRR's mass, diameter, drag coefficient, propellant mass fraction, exhaust velocity, initial inclination and propellant burning time.  Effective frontal area of the rocket is
      \begin{linenomath}\[
      A=\frac{\pi}{4}\ d^2.
      \]\end{linenomath}
      While the rocket engine is burning the thrust is obtained by combining (\ref{2.02}) and (\ref{2.03}):
      \begin{linenomath}\[
      F_t=\frac{M\ f_p}{t_b}\ v_e,
      \]\end{linenomath}
      where $M$ is the rocket mass, $f_p$ is the propellant mass fraction, $t_b$ is the engine burning time, and $v_e$ is the exhaust velocity.
      Using this data, the program starts to perform iterative calculations on the rocket.

      \hskip0.8cm At time $t=0$, Rocket.m has the following:
      rocket's mass $M$;
      rocket's coordinates in $xy$ -- plane, where $x$ is the horizontal distance and $y$ is the altitude;
      rocket's velocity $\mathbf{v}$ -- which has horizontal component $v_x$ and vertical component $v_y$.
      Rocket.m uses the rocket's altitude to determine air density $\rho$ and sonic velocity $v_s$ at time $t=0$.  For air density as a function of altitude, we use US Standard Atmosphere 1976. Rocket.m calculates the Mach number as $\mathcal{M}=v/v_s$.  The drag coefficient $C_d$ is determined from the Mach number.  The drag force is given by by (\ref{2.01}):
      \begin{linenomath} \be
        \label{A.01}
        F_d=C_d(\mathcal{M})\ \ \frac{\rho v^2 A}{2}.
        \ee \end{linenomath}
      The drag force acts in the direction opposite of rocket motion.  After the rocket's engine burns out, $F_d=0$.  Rocket acceleration is derived from all forces acting on the rocket as shown in Figure \ref{F03}:
      \begin{linenomath} \be
        \label{A.02}
        \mathbf{a}=\frac{F_t-F_d}{M(t)}\ \hat{v}-g\ \hat{y},
      \ee \end{linenomath}
      where $\hat{v}$ is a unit vector in the direction of rocket motion, and $\hat{y}$ is a unit vector in $y$ or upward direction.

      \hskip0.8cm Rocket.m uses the information given in the above paragraph to calculate similar information at a time $dt$.  If the rocket's engines are still burning, then the rocket's mass has decreased by the mass of propellent consumed:
      \begin{linenomath}\[
      dM=\frac{M\ f_p}{t_b}\ dt,
      \]\end{linenomath}
      otherwise the rocket mass stays constant.
      The rocket's velocity has changed by
      \begin{linenomath}\[
      \mathbf{v(t+dt)} = \mathbf{v(t)}+\mathbf{a(t)}\ dt.
      \]\end{linenomath}
      The rocket's position has changed by
      \begin{linenomath}\[
      \mathbf{r(t+dt)} = \mathbf{r(t)}+\mathbf{v(t)}\ dt.
      \]\end{linenomath}
      Based on the new mass, position, and velocity,
      Rocket.m calculates the rocket's drag force and acceleration at time $dt$.

      \hskip0.8cm Rocket.m performs another iteration obtaining the rocket's drag force and acceleration at time $2\ dt$.  Rocket.m performs many iterations, until the rocket reaches its apogee -- the maximal ascent.  Generally, $dt=0.1\ s$ and it takes 140 $s$ to 160 $s$ for the rocket to reach the apogee.  Rocket.m performs 1,400 to 1,600 iterations within a few seconds.

      \hskip0.8cm Rocket.m calculates DLSRR's flight apogee altitude $y$, flight apogee horizontal velocity $v_x$, and flight apogee horizontal coordinate $x$.  Recall, that apogee vertical velocity is $v_y=0$.  Rocket.m also calculates DLSRR's drag loss by Eq. (\ref{2.07}) and gravity loss by Eq. (\ref{2.12}).

      \hskip0.8cm Rocket.m produces a time series of DLSRR's coordinates $(x,y)$, and velocity
      $\mathbf{v}=\big( v_x,v_y \big)$ with time interval $dt$.  The time series of ambient air densities and temperatures determined by altitude $y$ as well as the Mach numbers are also recorded.

  \item[Impact.m] performs calculations on the projectiles released at the apogee.  The user inputs the projectile's mass, diameter, drag coefficient, apogee altitude, apogee, velocity, and apogee horizontal coordinate.  The program calculates impact distance and velocity.

      \hskip0.8cm Mathematically, Impact.m is a simplified version of Rocket.m.  As the rocket reaches the apogee, it releases one or more warhead.  The warhead(s) keep the rocket's horizontal and vertical coordinates $(x,y)$.  They keep the rocket's horizontal velocity $v_x$, and vertical velocity, which is $0$ at the apogee.  The warhead(s) have a new caliber $d_1 \ll d$, and hence a new frontal area
      \begin{linenomath}\[
      A_1=\frac{\pi}{4}\ d_1^2.
      \]\end{linenomath}
      They also have a new mass $M_1 \ll M$, which stays the same throughout the descent.
      The warhead(s) have new drag coefficient dependance on Mach number $C_{d1}(\mathcal{M})$.

      \hskip0.8cm As the warhead(s) descend from the apogee to the target, Impact.m calculates their acceleration, velocity, and coordinate for each time increment $dt$.  Drag force is calculated by Eq. (\ref{A.01}) with the new frontal area and drag coefficient function.  Acceleration is calculated by
      \begin{linenomath} \be
        \label{A.03}
        \mathbf{a}=-\frac{F_d}{M(t)}\ \hat{v}-g\ \hat{y},
      \ee \end{linenomath}
      where $\hat{v}$ is a unit vector in the direction of rocket motion, and $\hat{y}$ is a unit vector in $y$ or upward direction.

      \hskip0.8cm Impact.m calculates warhead's descent time, impact velocity, and horizontal distance of the impact.  Impact.m also calculates the warhead's drag loss by Eq. (\ref{2.07}).  The warhead experiences no gravity loss during the descent.  The combination of Rocket.m and Impact.m calculate the warheads horizontal range from the point of rocket release, and the flight time from rocket firing to impact.

  \item[ThermalAnalysis.m] calculates the temperatures of SSR stagnation point, several points on SSR nose cone, and rocket cylinder base during the powered portion of SSR ascent.  The program takes in the time series data containing ambient air temperature, ambient air density, and rocket velocity.  These time series are produced by Rocket.m.

      \hskip0.8cm For every time interval $dt$, ThermalAnalysis.m solves Eq. (\ref{5.31}) for the stagnation point wall temperature.  Likewise, for every time interval $dt$, ThermalAnalysis.m solves Eq. (\ref{5.33}) for wall temperature at several points on the rocket's nose.  Notice, that the temperatures are found for each time independently of temperatures in the previous times.

      \hskip0.8cm ThermalAnalysis.m uses Eq.(\ref{5.23}) to calculate the heat flux on the rocket cylinder base caused by the turbulent flow.  The cylinder does not radiate away any of absorbed heat.  Being a heat sink shield, it absorbs the heat and gets hotter.
      ThermalAnalysis.m assumes that the rocket cylinder base starts at 20 $^o$C, and calculates the time series of rocket base temperatures as the rocket base absorbs heat flux.  ThermalAnalysis.m displays the cylinder base temperature at the end of rocket firing.

      \hskip0.8cm Rocket wall temperature time series at stagnation point as well as distances of 3 $cm$, 10 $cm$, and 30 $cm$ from the stagnation point are plotted in Figure \ref{F11}.  Cylinder base temperature at the end of rocket firing are presented in Table \ref{T10}.

\end{description}


\end{document}